\pdfoutput=1 
\documentclass[12pt]{report}   %12 point font for Times New Roman
\usepackage{graphicx}  %for images and plots
\usepackage[letterpaper, left=1in, right=1in, top=1in, bottom=1in]{geometry}
\usepackage{setspace}  %use this package to set linespacing as desired
\usepackage{times}  %set Times New Roman as the font
\usepackage[explicit]{titlesec}  %title control and formatting
\usepackage[titles]{tocloft}  %table of contents control and formatting
\usepackage[backend=bibtex, sorting=none, bibstyle=ieee]{biblatex}  %reference manager
\usepackage[bookmarks=true, hidelinks]{hyperref}
\usepackage[page]{appendix}  %for appendices
\usepackage{rotating}  %for rotated, landscape images
\usepackage[normalem]{ulem}  %for italicized text
\usepackage{amsmath}
\newcommand{\etal}{et al.}
\DeclareMathOperator*{\argmin}{arg\,min} 
\bibliography{references}

\AtEveryBibitem{\clearfield{issn}}
\AtEveryBibitem{\clearlist{issn}}

\AtEveryBibitem{\clearfield{language}}
\AtEveryBibitem{\clearlist{language}}

\AtEveryBibitem{\clearfield{doi}}
\AtEveryBibitem{\clearlist{doi}}

\AtEveryBibitem{\clearfield{url}}
\AtEveryBibitem{\clearlist{url}}

\AtEveryBibitem{%
  \ifentrytype{online}
    {}
    {\clearfield{urlyear}\clearfield{urlmonth}\clearfield{urlday}}}

\begin{document}
\doublespacing  %set line spacing

%%%%%%%%%%%%%%%%%%%%%%%%%%%%%%%%%%%%%
% Title Page
%%%%%%%%%%%%%%%%%%%%%%%%%%%%%%%%%%%%%

%% Define your thesis title, your name, your school, and your month and year of graduation here

\newcommand{\thesisTitle}{Modeling, simulation, and optimization of a monopod hopping on yielding terrain}
\newcommand{\yourName}{Juntao He}
\newcommand{\yourSchool}{Fiction}
\newcommand{\yourMonth}{August}
\newcommand{\yourYear}{2021}

%%%%%%%%%%%%%%%%%%%%%%%%%%%%%%%%%%%%%%%%%%%%%%%%%%%%%%%%%
% Do not edit these lines unless you wish to customize
% the template
%%%%%%%%%%%%%%%%%%%%%%%%%%%%%%%%%%%%%%%%%%%%%%%%%%%%%%%%%

\begin{titlepage}
\begin{center}

\begin{singlespacing}

\vspace{15\baselineskip}
\textbf{\MakeUppercase{\thesisTitle}}\\

\vspace{3\baselineskip}
By\\
\vspace{3\baselineskip}
\yourName\\
\vspace{2\baselineskip}
Master's thesis\\
\vspace{2\baselineskip}
Submitted in Partial Fulfillment of the\\
Requirements for the Degree of\\
\vspace{3\baselineskip}
MASTER OF SCIENCE IN MECHANICAL ENGINEERING\\
\vspace{3\baselineskip}
EVANSTON, Illinios\\
\vspace{1\baselineskip}
\yourMonth{} \yourYear{}\\

\vfill

\end{singlespacing}

\end{center}
\end{titlepage}

\currentpdfbookmark{Title Page}{titlePage}  %add PDF bookmark for this page

%%%%%%%%%%%%%%%%%%%%%%%%%%%%%%%%%%%%%
%copyright page
%%%%%%%%%%%%%%%%%%%%%%%%%%%%%%%%%%%%%
%\input{copyright.tex}

%%%%%%%%%%%%%%%%%%%%%%%%%%%%%%%%%%%%%%%%%%%%%%%%%%%%%%%%%%%%%%%%%
% This is the abstract 
%%%%%%%%%%%%%%%%%%%%%%%%%%%%%%%%%%%%%%%%%%%%%%%%%%%%%%%%%%%%%%%%%
%\pagenumbering{arabic}
%\setcounter{page}{2} % set the page number appropriately

%\input{abstract.tex}

%%%%%%%%%%%%%%%%%%%%%%%%%%%%%%%%%%%%%
% Acknowledgments
%%%%%%%%%%%%%%%%%%%%%%%%%%%%%%%%%%%%%

\addcontentsline{toc}{chapter}{Acknowledgments}

\clearpage
\begin{centering}
\textbf{ACKNOWLEDGEMENTS}\\
\vspace{\baselineskip}
\end{centering}

%Insert your acknowledgements text here
First of all, I would like to thank Prof. Kevin Lynch, Prof. Paul Umbanhowar and Dan Lynch for their invaluable help and support on my master's thesis.

I would also like to thank all the staff at Northwestern University, especially those who fight on the Covid-19 front line. Without them, I could never finish my thesis and course of study in these tough times.  

Finally, I would like to thank my parents for their support of my master's study. They are my strong backing in the Covid pandemic. 

\clearpage
%\pagenumbering{gobble}  %remove page number on summary page

%\addtocontents{toc}{\cftpagenumbersoff{chapter}} 

%\currentpdfbookmark{Acknowledgments}{acknowledgments}
%\addtocontents{toc}{\cftpagenumberson{chapter}} 
%%%%%%%%%%%%%%%%%%%%%%%%%%%%%%%%%%%%%
% Preface
%%%%%%%%%%%%%%%%%%%%%%%%%%%%%%%%%%%%%

%\input{preface.tex}

%%%%%%%%%%%%%%%%%%%%%%%%%%%%%%%%%%%%%
% Abbreviations
%%%%%%%%%%%%%%%%%%%%%%%%%%%%%%%%%%%%%

%\input{abbreviations.tex}

%%%%%%%%%%%%%%%%%%%%%%%%%%%%%%%%%%%%%
% Glossary
%%%%%%%%%%%%%%%%%%%%%%%%%%%%%%%%%%%%%

%\input{glossary.tex}

%%%%%%%%%%%%%%%%%%%%%%%%%%%%%%%%%%%%%
% Nomenclature
%%%%%%%%%%%%%%%%%%%%%%%%%%%%%%%%%%%%%

%\input{nomenclature.tex}

%%%%%%%%%%%%%%%%%%%%%%%%%%%%%%%%%%%%%
% Table of Contents
%%%%%%%%%%%%%%%%%%%%%%%%%%%%%%%%%%%%%

% Format for Table of Contents
\renewcommand{\cftchapdotsep}{\cftdotsep}  %add dot separators
\renewcommand{\cftchapfont}{\bfseries}  %set title font weight
\renewcommand{\cftchappagefont}{}  %set page number font weight
\renewcommand{\cftchappresnum}{Chapter }
\renewcommand{\cftchapaftersnum}{:}
\renewcommand{\cftchapnumwidth}{5em}
\renewcommand{\cftchapafterpnum}{\vskip\baselineskip} %set correct spacing for entries in single space environment
\renewcommand{\cftsecafterpnum}{\vskip\baselineskip}  %set correct spacing for entries in single space environment
\renewcommand{\cftsubsecafterpnum}{\vskip\baselineskip} %set correct spacing for entries in single space environment
\renewcommand{\cftsubsubsecafterpnum}{\vskip\baselineskip} %set correct spacing for entries in single space environment

%format title font size and position (this also applys to list of figures and list of tables)
\titleformat{\chapter}[display]
{\normalfont\bfseries\filcenter}{\chaptertitlename\ \thechapter}{0pt}{\MakeUppercase{#1}}

\renewcommand\contentsname{Table of Contents}

\begin{singlespace}
\tableofcontents
\end{singlespace}

\currentpdfbookmark{Table of Contents}{TOC}

\clearpage

%%%%%%%%%%%%%%%%%%%%%%%%%%%%%%%%%%%%%
% List of figures and tables
%%%%%%%%%%%%%%%%%%%%%%%%%%%%%%%%%%%%%

\addcontentsline{toc}{chapter}{List of Tables}
\begin{singlespace}
	\setlength\cftbeforetabskip{\baselineskip}  %manually set spacing between entries
	\listoftables
\end{singlespace}

\clearpage

\addcontentsline{toc}{chapter}{List of Figures}
\begin{singlespace}
\setlength\cftbeforefigskip{\baselineskip}  %manually set spacing between entries
\listoffigures
\end{singlespace}

\clearpage

%%%%%%%%%%%%%%%%%%%%%%%%%%%%
% CHAPTERS
%%%%%%%%%%%%%%%%%%%%%%%%%%%%

%%%%%%%%%%%%%%%%%%%%%%
% formatting
%%%%%%%%%%%%%%%%%%%%%%

% resume page numbering for rest of document
\clearpage

% Adjust chapter title formatting
\titleformat{\chapter}[display]
{\normalfont\bfseries\filcenter}{\MakeUppercase\chaptertitlename\ \thechapter}{0pt}{\MakeUppercase{#1}}  %spacing between titles
\titlespacing*{\chapter}
  {0pt}{0pt}{30pt}	%controls vertical margins on title
  
% Adjust section title formatting
\titleformat{\section}{\normalfont\bfseries}{\thesection}{1em}{#1}

% Adjust subsection title formatting
\titleformat{\subsection}{\normalfont}{\uline{\thesubsection}}{0em}{\uline{\hspace{1em}#1}}

% Adjust subsubsection title formatting
\titleformat{\subsubsection}{\normalfont\itshape}{\thesubsection}{1em}{#1}

%%%%%%%%%%%%%%%%
% Introduction
%%%%%%%%%%%%%%%%

%\input{introduction.tex}

%%%%%%%%%%%%%%%%
% Chapter 1
%%%%%%%%%%%%%%%%

\chapter{Introduction and Background}
Mobile robots have great potential to improve search and rescue, disaster response, factory inspection and package delivery. Legged locomotion has significant advantages over wheeled locomotion on yielding surfaces, such as soil, sand, snow and gravel. However, researchers have mainly focussed on generating stable walking and running gaits on rigid flat
\cite{Grizzle_bipedal_walker_runner} or uneven  terrain \cite{Walking_over_rough_terrain}. There are far fewer studies about legged locomotion on yielding ground.\newline 
Researchers already made robots to accomplish some locomotion tasks on soft ground. Earlier work used fixed kinematics (fixed gaits) to move on soft ground \cite{Li_Terradynamics_2013} \cite{Aguilar_lift_off_robot}. Only recently has the community started to look at feedforward control \cite{terrain_aware_motion_planning}. In \cite{terrain_aware_motion_planning}, Hubicki \etal \text{ }present  a  model  and  fast  optimization  formulation which  generates  accurate  motion  plans  on  granular  media  with relatively short solving  times. Reference \cite{xiong_stability_IROS_2017} achieved flat-footed  bipedal  walking  on deformable  granular  terrain with feedback control, but they did so by seeking to avoid ground penetration. Reference \cite{roberts2018reactive} suggests an active damping controller to reduce the energy cost for monopedal vertical hopping on soft ground, while reference \cite{soft_landing} seeks to minimize energy loss on soft-landing using feedforward force control as well as optimal impedance.\newline
In this work, I use feedforward control to develop a periodic gait on soft ground using a ground reaction force model. I test this approach by performing discrete element method (DEM) simulations of the monoped robot hopping on granular materials. To stabilize the target gait, I introduce a feedback term.

\section{Chrono simulation environment}
To test the force control approach I develop for periodic hopping on soft ground, experimental validation using a real robot on real ground would be preferable. However, due to the challenges of working in the lab during the Covid-19 pandemic, I instead chose to validate the control approach in a virtual environment, namely Chrono. Chrono\footnote{\url{http://api.projectchrono.org/5.0.0/}} is a physics engine with specialized support for simulating interaction between rigid bodies and granular media. Chrono supports C++ and Python, but the support for Python is still under construction, so to take full advantage of Chrono's features, C++ is recommended. Chrono’s discrete element method (DEM) code works with a GPU but there is no need for users to write GPU code because Chrono’s developers already provided libraries for the GPU part. Chrono provides interfaces with CAD programs like SOLIDWORKS to simplify model building in Chrono. Chrono also has better integration of rigid-body dynamics with DEM simulation than other DEM simulators such as LAMMPS\footnote{\url{https://lammps.sandia.gov/}} and LIGGGHTS\footnote{\url{https://www.cfdem.com/liggghts-open-source-discrete-element-method-particle-simulation-code}}. \newline 
In the following chapters, I use Chrono to generate resistive forces approximation functions of the ground reaction force and test the analytic feedforward force control solution generated by MATLAB using the RFT model. Advantages of the Chrono simulation environment over real world experiments include:
\begin{enumerate}
    \item To the extent that the DEM simulation is physically accurate, in-simulation validation provides more feedback than hardware validation, since potentially any variable of interest can be measured and recorded. This results in a better-informed motion planning framework.
    \item The cost of experimental setup and the cost of experimental failure are both lower than with hardware experiments.
    \item The performance of the robot and its control on different soft substrates can be investigated by simply changing the parameters related to the physical properties of the granular materials.
\end{enumerate}

\section{Thesis outline}

The structure of the thesis is as follows: \newline
Chapter \ref{chapter 2} validates the reliability and accuracy of the Chrono simulation environment by comparing simulation results with experimental results from Li \etal \textrm{} \cite{Li_Terradynamics_2013}. Chapter \ref{Chapter 3} generates functions to predict ground reaction forces from granular media using a Fourier representation matched to Chrono simulation data. Chapter \ref{chapter 4} develops a control strategy which enables the 1D hopping monopod to jump back to its original maximum height. Chapter 5 compares the performance of feedforward controls in the DEM simulation for a range of hop heights and for hoppers with different foot areas. Chapter 6 presents the conclusions.

%%%%%%%%%%%%%%%%
% Chapter 2
%%%%%%%%%%%%%%%%

\chapter{Chrono environment validation} \label{chapter 2}
To validate the Chrono simulation environment, I compared my simulation results with the experimental results in Li \etal \textrm{} \cite{Li_Terradynamics_2013}.\newline
Refrence \cite{Li_Terradynamics_2013} finds that the granular resistive stresses are linear with depth, but only for low intrusion speeds (less than 0.5 m/s), only in the bulk, i.e., away from the container sides (to avoid Janssen effects \cite{Janssen}) and at depths both large enough to avoid nonlinearities due to free particles at the surface and small enough to avoid nonlinearities due to compression against the container bottom. For a rigid plate intruder moving in granular media \cite{Li_Terradynamics_2013}, the lithostatic stresses can be modeled as
\begin{equation}
\label{stress}
\sigma_{z,x}(|z|,\beta,\gamma) =  \left\{
\begin{array}{rl}
\alpha_{z,x}(\beta,\gamma)|z| & z < 0,\\
0 &  z > 0,
\end{array} \right.
\end{equation}
where $\sigma_{z,x}$ are the vertical ($z$-component) and horizontal ($x$-component) resistive stresses, $z$ is the depth of the intruder, and $\alpha_{z,x}$ are vertical and horizontal stresses per unit depth in the linear response regime. Note that when the intruder is above the granular domain ($z > 0$), there are no resistive stresses. Angle of attack $\beta$ and angle of intrusion $\gamma$ are shown in Figure \ref{fig:penetration}.
\section{Simulation setup} \label{sim_setup_1}
\begin{figure}
    \centering
    \includegraphics[width=13cm]{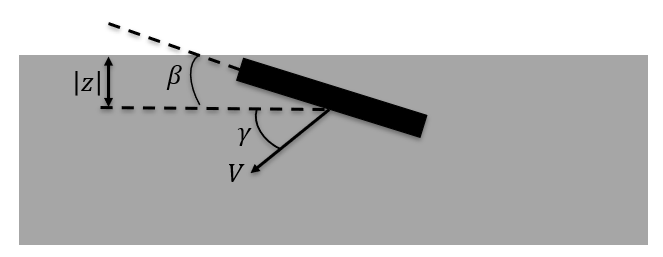}
    \caption[Sketch of a rigid plate intruder]{ The rigid plate intrudes into the granular media with velocity $V$, attack angle $\beta$ and intrusion angle $\gamma$. The distance between the CoM of the plate and the granular bed top, $\mid z \mid$, is the intrusion depth. }
    \label{fig:penetration}
\end{figure}

Figure \ref{fig:penetration} shows a planar view of the plate penetrating into the granular domain. The black rectangle represents the plate while the colored area represents the granular media. The plate has dimensions of 3.81 cm $\times$ 2.54 cm $\times$ 0.64 cm (area = 9.68 $\textrm{cm}^2$) and a density of 2.7 $\mbox{g/cm}^3$, which match the dimensions and density of the flat plate used by Li et al. in \cite{Li_Terradynamics_2013}. All the spheres in the granular domain have identical physical properties, specifically a radius of 3.2 mm and a density of 2.6 $\mbox{g/cm}^3$. The spheres are contained in a box with cross section 24 cm $\times$ 22 cm, and fill the box to a depth of approximately 18 cm. The bed is prepared by filling the container from the bottom to the top layer by layer. Li et al. used a few different granular materials in their experiments. I chose 3.2 mm diameter glass spheres instead of those materials with small radius here because I can use far fewer granular spheres for my simulation than smaller radius. This makes our simulation time cost more acceptable. That is, the simulation time for a specific $\gamma-\beta$ pair intrusion trial is 2 to 3 minutes. For materials with larger radii, fewer particles contact a fixed size plate, resulting in larger force fluctuations \cite{Miyai_particle_size_influence}, a topic I return to in the context of feedback control in Chapter \ref{chapter 4}. Considering both reducing force fluctuations and simulation time, 3.2 mm is a good choice for the granular particle radius.  Table \ref{table:1} lists the simulation parameters, for more details, refer to the JSON file in my GitHub repository \footnote{\url{https://github.com/HappyLamb123/Foot-ROBOT/blob/master/demo_code/plate/demo_GRAN_plate.json}}. The friction is chosen the same as used in \cite{Li_Terradynamics_2013}. Damping coefficients are not mentioned in \cite{Li_Terradynamics_2013}, I set them as reasonable values for glass. \newline
\begin{table}[]
\begin{tabular}{ll}
\hline
Time step                                    & 1e-5 s                      \\
Plate dimension                              & 3.81 cm $\times$ 2.54 cm $\times$ 0.64 cm   \\
Plate density                                & 2.7 $\mbox{g/cm}^3$ \\
Sphere radius                     & 3.2 mm                      \\
Sphere density                    & 2.6 $\mbox{g/cm}^3$ \\
Domain size                    & 24 cm $\times$ 22 cm $\times$ 18 cm (depth)         \\
Sphere-sphere normal contact stiffness    & 1e8 N/m                    \\
Sphere-wall normal contact stiffness      & 1e8 N/m                      \\
Sphere-sphere normal damping coefficient  & 6000 N s/m                 \\
Sphere-wall normal damping coefficient    & 6000 N s/m                \\
Sphere-sphere tangent contact stiffness   & 1e6 N/m                    \\
Sphere-wall tangent contact stiffness     & 1e6 N/m                    \\
Sphere-sphere tangent damping coefficient & 0                          \\
Sphere-wall tangent damping coefficient   & 0                          \\
Sphere-sphere static friction coefficient & 0.63                       \\
Sphere-wall static friction coefficient   & 0.63                       \\ \hline
\end{tabular}
\caption[Chrono simulation parameters]{Parameters for Chrono granular simulations.}
\label{table:1}
\end{table}\newline
To reproduce the stress parameters results for 3 mm glass spheres from the experiments shown in Fig. S4(I, J) in the supplementary materials of \cite{Li_Terradynamics_2013}, I did 271 simulations with attack angle $\beta$ and intrusion angle $\gamma$ varied from -90 degrees to 90 degrees (three repeated trials for each $\beta$ and $\gamma$ pair). For the intrusion speed $V$, I chose 3 cm/s because it guarantees both accuracy(how close the simulation results are to experimental results in \cite{Li_Terradynamics_2013}) and time-efficiency. Further discussion of the choice of intrusion speed can be found at the end of this chapter. In each simulation, $\beta$ and $\gamma$ are fixed during penetration and the absolute value of intrusion speed $V$ is 3 cm/s. In the Chrono simulation, I record the horizontal and vertical resistive forces of the granular media every microsecond into a CSV file. Then, I divide the resistive forces by the area of the rigid plate to get the stress in the $x$- and $z$- direction. To get the stress gradient, I linearly fit the horizontal and vertical direction stress from 2 cm to 7.5 cm. The resulting slopes of the lines are the stress gradient, $\alpha_{z,x}$. I choose the depth range for fitting in order to avoid surface effects and effects from the container bottom.
\section{Comparison between DEM results and experiments} \label{results compare}
\begin{figure}[htp]
    \centering
    \includegraphics[width=17cm]{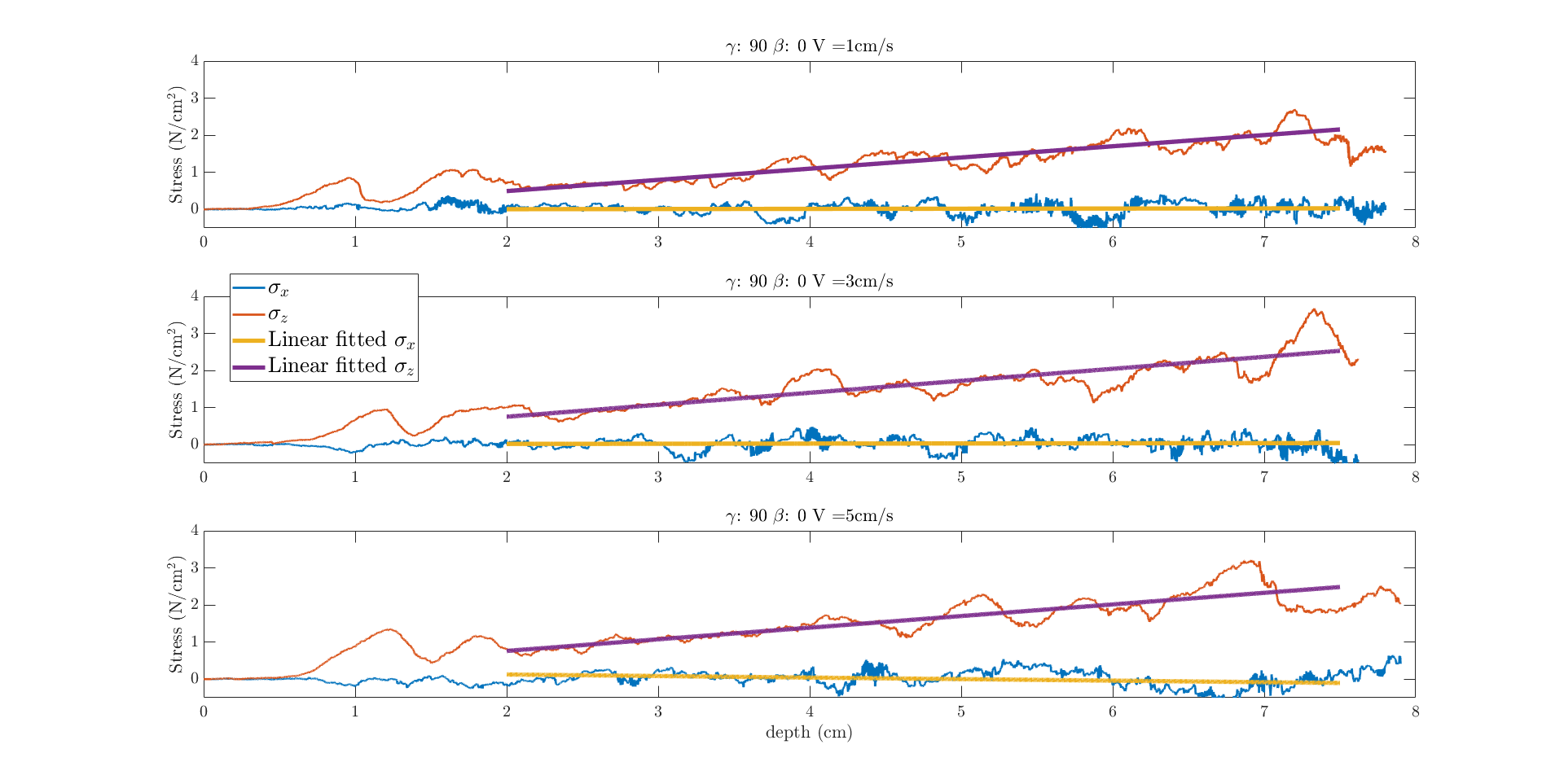}
    \caption[Stress results of three different penetration velocities]{Vertical ($z$) and horizontal ($x$) stresses vs. penetration depth and corresponding linear fits (lines). Similar results for intrusion velocities of 1, 3, and 5 cm/s indicate that 3 cm/s is a good choice for balancing simulation time and accuracy relative to experimental results in \cite{Li_Terradynamics_2013}. }
    \label{fig:speed}
\end{figure}

\begin{figure}
    \centering
    \includegraphics[width=15cm]{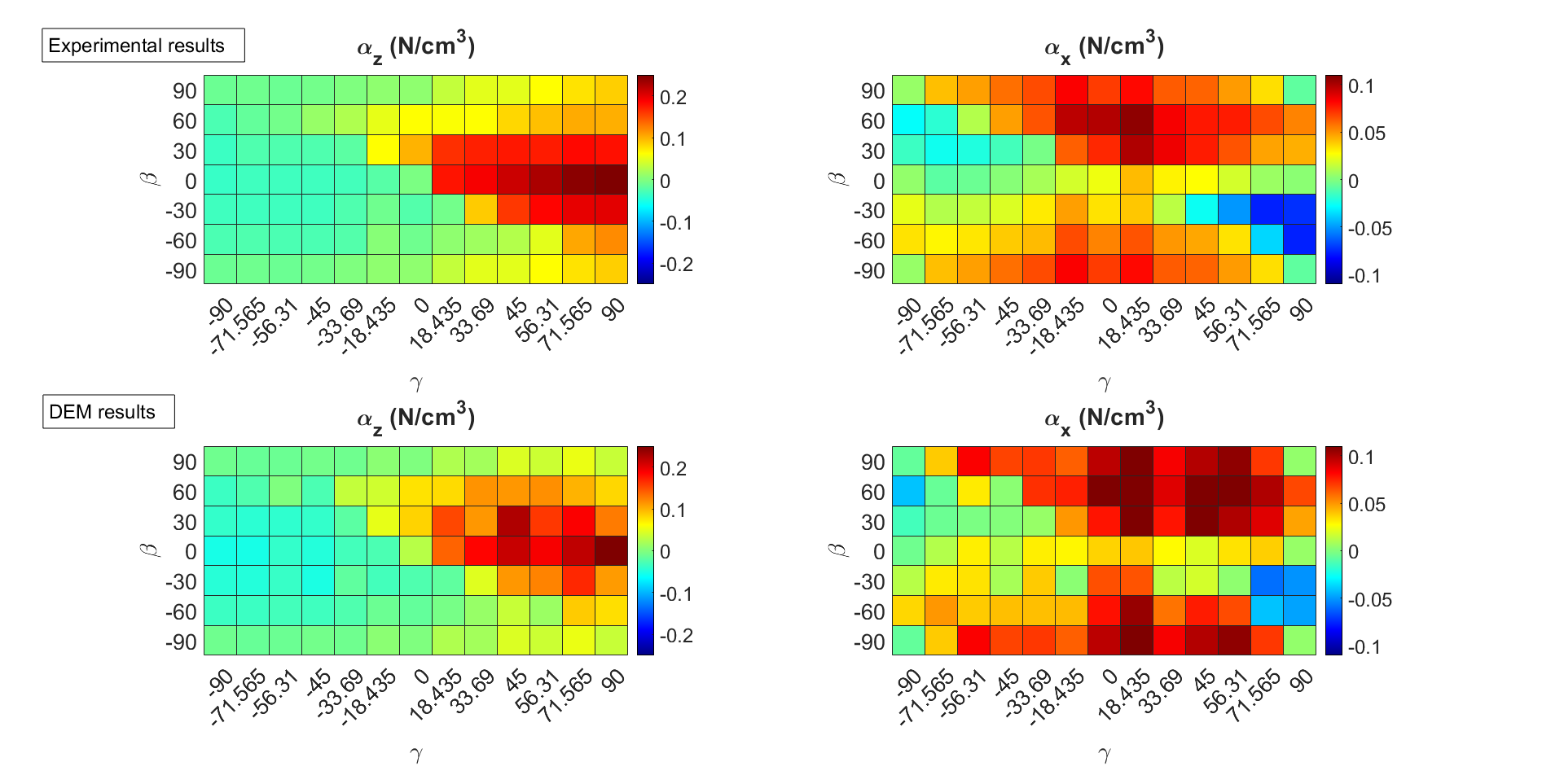}
    \caption[Comparison of experimental and DEM results]{Experimental and DEM results for vertical and horizontal stresses indicate good quantitative agreement for all $\gamma$ and $\beta$.}
    \label{fig:combinel}
\end{figure}

\begin{figure}
    \centering
    \includegraphics[width=15cm]{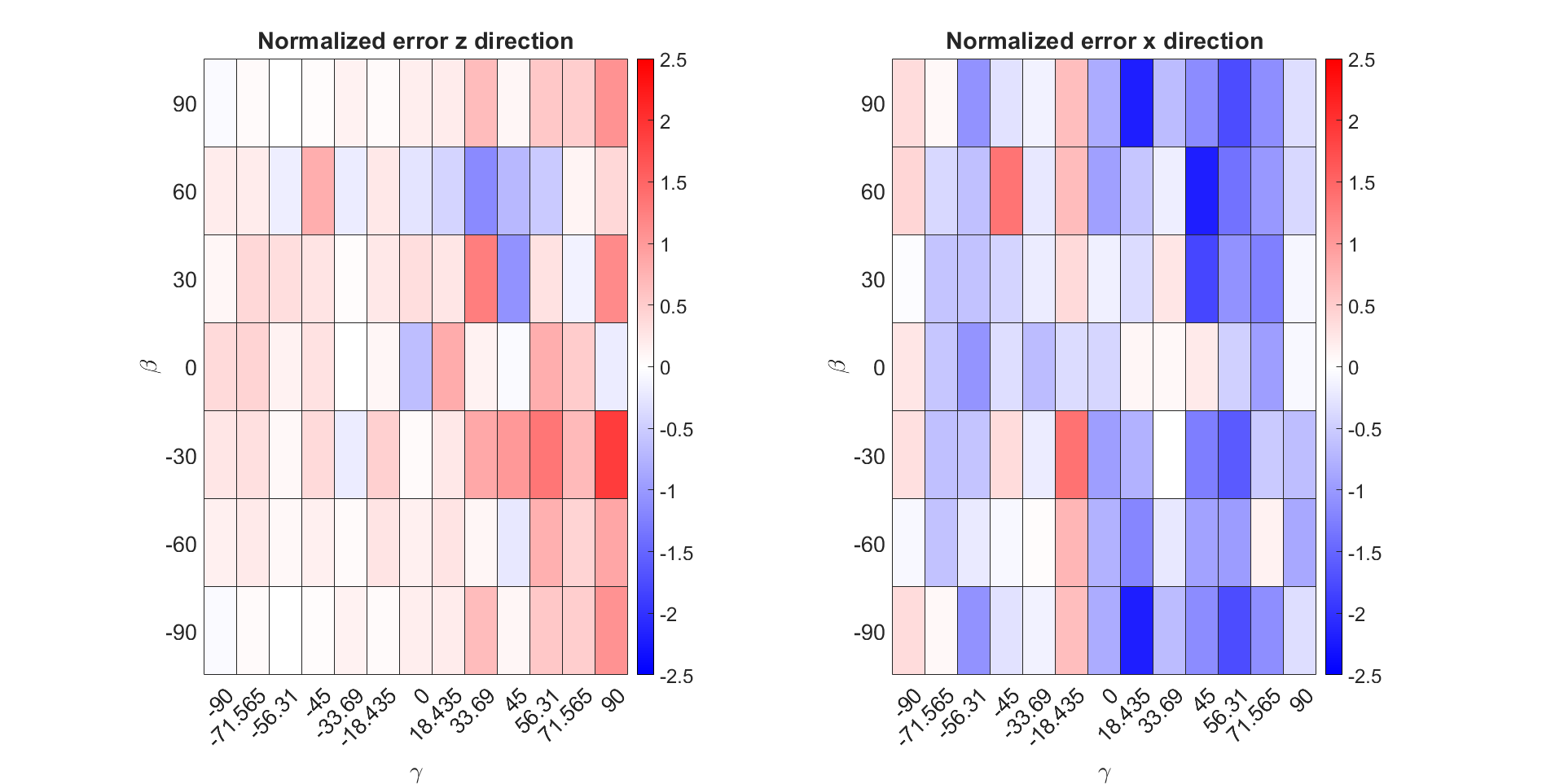}
    \caption[Normalized error map]{Normalized error map (see equation \ref{eq1}).}
    \label{fig:error}
\end{figure}

As \cite{Li_Terradynamics_2013} suggests, the stresses in the horizontal and vertical directions are proportional to the penetration depth as Figure \ref{fig:speed} shows. By linearly fitting the stress versus depth curves in MATLAB, the stress gradients ($\alpha_x$ and $\alpha_z$) can be calculated as the slopes of those fitted curves.
Figure \ref{fig:combinel} shows the Chrono simulation results and experimental results of $x$ and $z$ direction stress gradients for different $\beta$ and $\gamma$. These DEM simulation results show that horizontal and vertical resistive stresses scale linearly with intruder depth, indicating that the  DEM simulations are a viable alternative to experiments performed with real granular media. The upper two plots in Figure \ref{fig:combinel} show experimental results from Li \etal , while the lower two are DEM results generated by Chrono. As Figure \ref{fig:combinel} indicates, the simulation results closely match the experimental results. To better characterize the accuracy of the Chrono simulation environment, a normalized error map is presented in Figure \ref{fig:error}.  The normalized errors in Figure \ref{fig:error}. are calculated as,\newline
\begin{equation}
\label{eq1}
\varepsilon_i=\frac{\alpha_{i, \mathrm{DEM}}-\alpha_{i,\mathrm{Exp}}}{\bar{\alpha}_{i,\mathrm{Exp}}},
\end{equation}
where $i$ is $x$ or $z$, $\alpha_{i,\mathrm{DEM}}$ represents the horizontal ($x$) or vertical ($z$) direction stress per cubic centimeter of DEM results while $\alpha_{i,\mathrm{Exp}}$ represents experimental results of Li \etal\cite{Li_Terradynamics_2013}. The $\bar{\alpha}_{i,\mathrm{Exp}}$ term in the denominator represents the mean value of all the experimental results .\newline
As the color map in Figure \ref{fig:combinel} indicates, the normalized errors are close to zero in most regions of the $\beta-\gamma$ plane. Only a few parts are dark red or blue where the simulation results and experimental results show larger differences. For $z$-direction normalized errors, the darkest color can be found in the column where $\gamma$ = 90 degrees. As for $x$-direction normalized errors, the darkest color can be found in the columns where $\gamma$ equals 18.435, 45 and 55.31 degrees. Other regions of the $\beta-\gamma$ plane have very small error values (less than 0.2). So the simulation results have a good match to the experimental results.
\newline
\section{The influence of the intrusion speed on ground reaction force}
In my simulations, the intrusion speed is 3 cm/s while Li et al. used 1 cm/s in their experiments. I opted for a faster intrusion speed because the rigid plate can reach the same depth in less time with a higher intrusion speed. A more acceptable time cost can help us to get more datasets in the same time period. For the robot control discussion in Chapter \ref{chapter 4}, a more efficient simulation time can help reduce the barrier to testing new control policies. With less time spent in simulation, I can iterate faster and converge to an effective control policy sooner. To determine the influence of speed variations, I ran three simulations with different penetration velocities (Figure \ref{fig:speed}). In those simulations, the simulation parameters are the same as I discussed above and the rigid plate moves downward in the vertical direction with a constant speed. The orientation of the plate was also fixed ($\beta$=0). Figure \ref{fig:speed} plots $\sigma_x$ and $\sigma_z$ versus penetration depth $z$ for three constant intrusion velocities $V$. Lines are fitted over a depth range of $2 \ \mathrm{ cm} \leq z \leq 7.5\ \mathrm{ cm}$. Table \ref{table:2} illustrates the actual mean values of three repeated trials of $\alpha_x$ and $\alpha_z$ for three velocities and the experimental results in Li’s paper \cite{Li_Terradynamics_2013} are added to the last row for comparison. As the table shows, 1 cm/s is within three percent of Li's results \cite{Li_Terradynamics_2013} for $\alpha_z$, but increasing the speed from 1 cm/s to 5 cm/s only changes $\alpha_z$ by seven percent. So, considering the time cost and the accuracy of my simulation, I choose 3 cm/s as the penetration speed for the DEM simulations.\newline

\begin{table}[]
\centering
\begin{tabular}{lll}
\hline
Impact velocity                    & $\alpha_x$ $[\mathrm{N}/\mathrm{cm}^3]$     & $\alpha_z$   $[\mathrm{N}/\mathrm{cm}^3]$ \\ \hline
Chrono (DEM) results       &           &        \\
1 cm/s               & 0.004631  & 0.3022 \\
3 cm/s             & 0.004155  & 0.2851  \\
5 cm/s                & -0.01086  & 0.3232  \\
Experimental results &           &        \\
1 cm/s              & 0.0025033 & 0.2931 \\ \hline
\end{tabular}
\caption[DEM stress gradient results of three different penetration velocities compared to experiments]{Vertically-constrained ($\beta = 0, \gamma = \pi/2$) penetration results for three different velocities. Variation of  $\alpha_x$ and $\alpha_z$ in three repeated trials is less than 0.003 $\mathrm{N}/\mathrm{cm}^3$ in all cases. }
\label{table:2}
\end{table}

%%%%%%%%%%%%%%%%
% Chapter 3
%%%%%%%%%%%%%%%%

\chapter{Ground Reaction Force model} \label{Chapter 3}
Modelling the response of yielding terrains to foot contact is challenging. The resistive forces from the ground are dependent on the granular compaction\cite{Umbanhowar_granular_impact} \cite{Qian_2015_principle_of_appendage_design}, the intruder kinematics (penetration depth and speed) \cite{Li_Terradynamics_2013} \cite{Aguilar2016RobophysicalSO} and the intruder morphology. Reference \cite{Aguilar2016RobophysicalSO} proposes that the resistive forces consist of a hydrodynamic-like term and a hydrostatic-like term, and investigates rapid intrusion by objects that change shape (self-deform) through passive and active means. \cite{Li_Terradynamics_2013} presents a granular resistive force theory to predict the forces on objects intruding relatively slowly(where inertial effects are negligible) with different directions and orientations.\newline
As discussed in the second paragraph of Chapter \ref{chapter 2}, \cite{Li_Terradynamics_2013} proposes that the granular resistive stresses are linear with depth. To obtain their stress results, Li \etal\cite{Li_Terradynamics_2013} intruded a rigid thin plate into a granular bed with different plate orientaions and movement directions. And then, they got stress gradients $\alpha_{z,x}(\beta,\gamma)$ (see Eq.\ref{stress}) by linearly fitting stress results at three different depth (2.54 cm, 5.08 cm, and 7.62 cm). Finally, they performed a discrete Fourier transform on $\alpha_{z,x}(\beta,\gamma)$ results over $-\pi/2 < \beta < \pi/2$ and  $-\pi < \gamma < \pi$ to obtain a fitting function. In this chapter, I do something similar but use more modes to get a more accurate representation. The fitting results in this Chapter will be used for further locomoting hopper simulations of our group.
\section{Simulation setup}
\begin{figure}
    \centering
    \includegraphics[width=17cm]{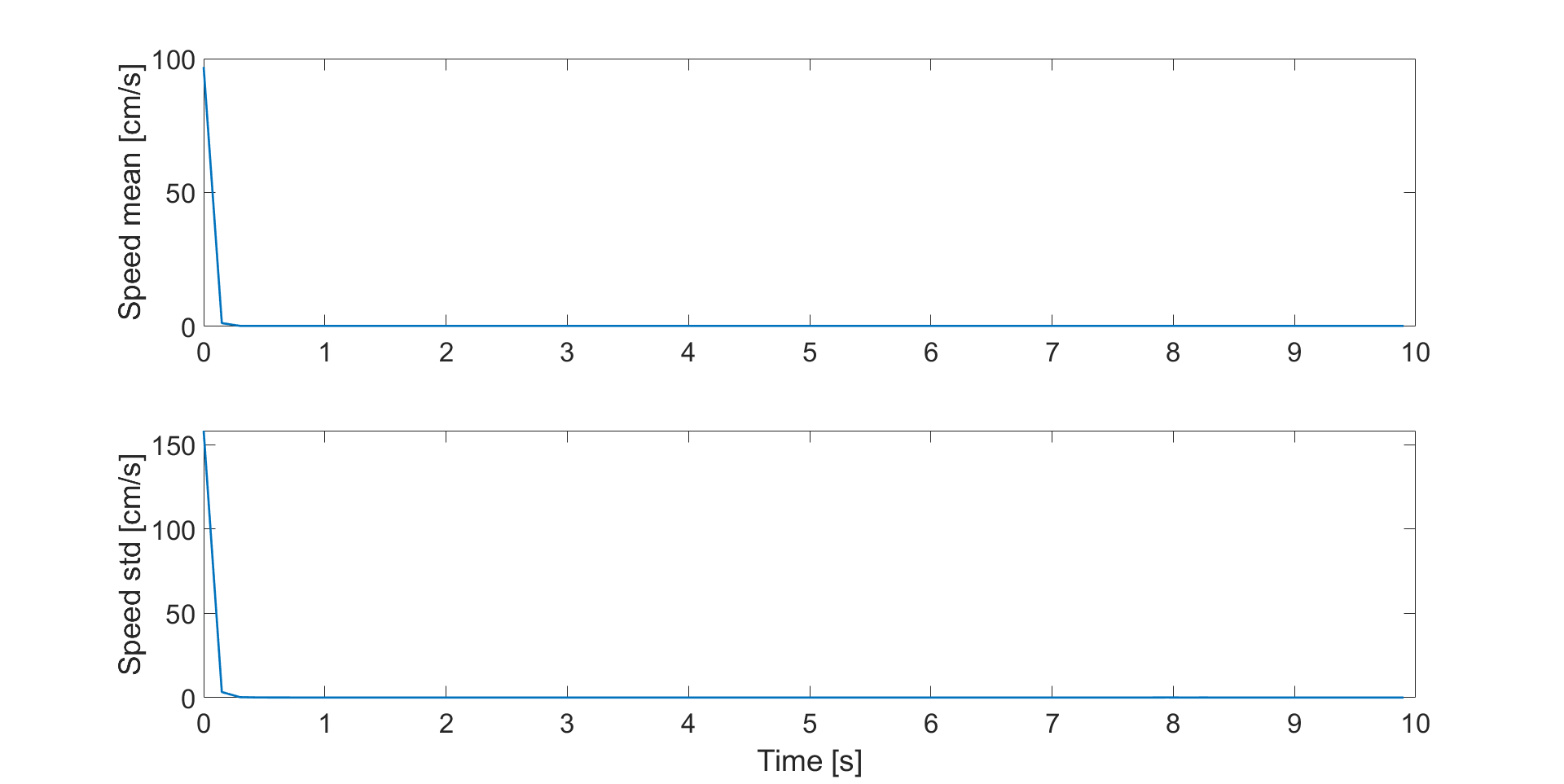}
    \caption[Bed cooling graph]{Average particle speed vs.\ time (upper) and particle speed standard deviation vs.\ time (bottom).}
    \label{fig:mean_std}
\end{figure}

\begin{figure}
    \centering
    \includegraphics[width = 17cm]{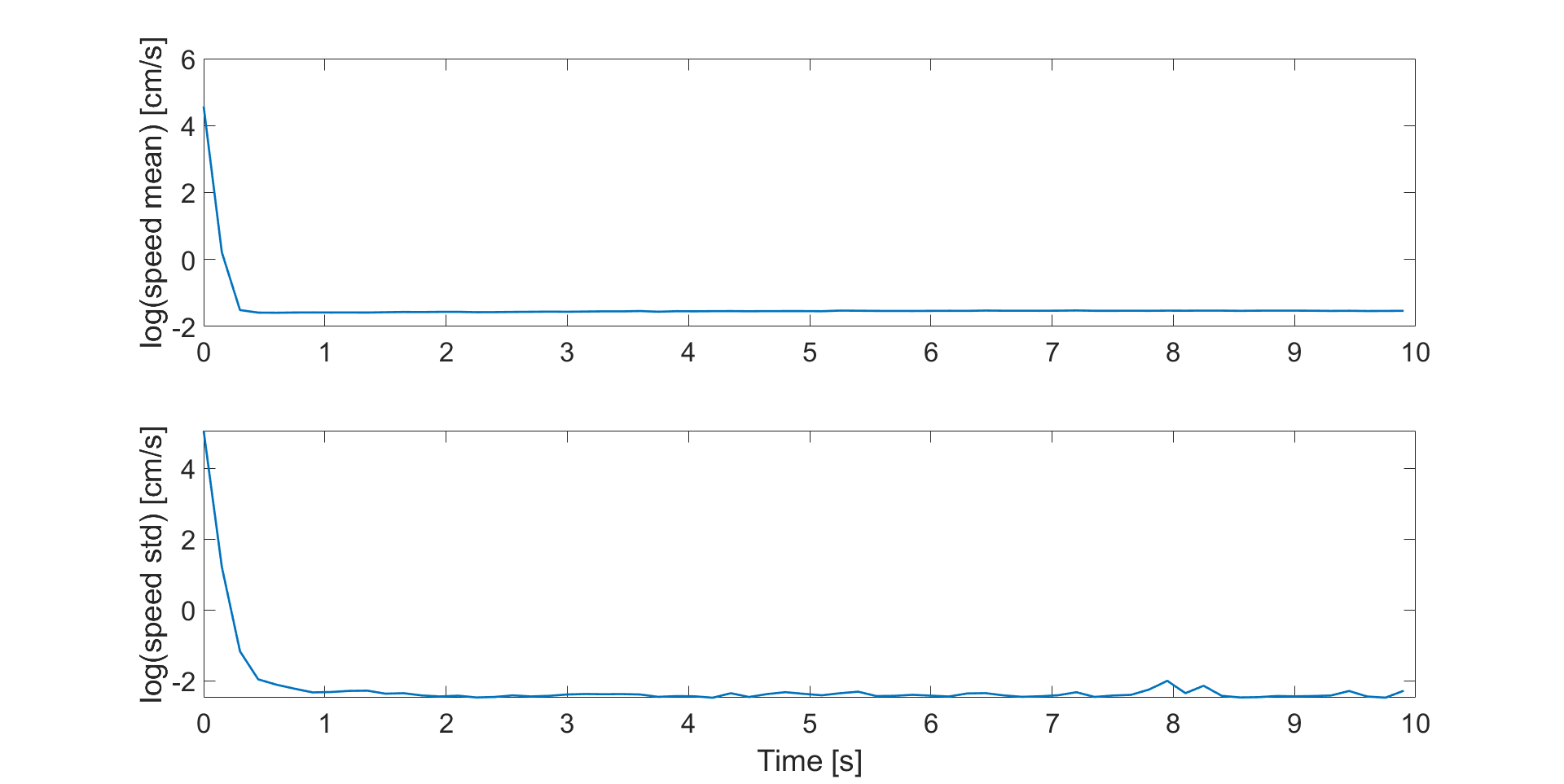}
    \caption[Logarithmic bed cooling graph]{Logarithmic average particle speed vs.\ time (upper) and logarithmic particle speed standard deviation vs.\ time (bottom).}
    \label{fig:log_mean_std}
\end{figure}

The simulation setup in this chapter is identical to in Figure \ref{fig:penetration}, which shows the planar view of the plate penetrating into the granular domain. The rigid plate is 5 cm $\times$ 5 cm $\times$ 0.5 cm (area = 25 $\textrm{cm}^2$) and the plate mass is 0.25 kg. More details about the simulation parameters can be found in Table \ref{table:2} and the JSON file in my GitHub repository \footnote{\url{https://github.com/HappyLamb123/Foot-ROBOT/blob/master/demo_code/generate_formula/intrude/demo_GRAN_plate.json}}.  
\begin{table}[]
\begin{tabular}{ll}
\hline
Time step                                    & 5e-5s                      \\
Plate dimension                              & 5 cm $\times$ 5 cm $\times$ 0.5 cm   \\
Plate density                                & 20 $\mbox{g/cm}^3$ \\
Sphere radius                     & 3 mm                      \\
Sphere density                    & 2.6 $\mbox{g/cm}^3$ \\
Domain size                    & 40 cm $\times$ 30 cm $\times$ 30 cm         \\
Sphere-sphere normal contact stiffness    & 1e8 N/m                    \\
Sphere-wall normal contact stiffness      & 1e8 N/m                      \\
Sphere-sphere normal damping coefficient  & 500 N s/m                 \\
Sphere-wall normal damping coefficient    & 500 N s/m                \\
Sphere-sphere tangent contact stiffness   & 1e8 N/m                    \\
Sphere-wall tangent contact stiffness     & 1e8 N/m                    \\
Sphere-sphere tangent damping coefficient & 500                          \\
Sphere-wall tangent damping coefficient   & 500                          \\
Sphere-sphere static friction coefficient & 0.385                       \\
Sphere-wall static friction coefficient   & 0.385                       \\ \hline
\end{tabular}
\caption[Chrono simulation parameters]{Parameters for Chrono granular simulations in Chapter \ref{Chapter 3}.}
\label{table:2}
\end{table}\newline
In Chapter \ref{chapter 2}, the granular particle container is filled by particles layer by layer from bottom to top, and no further action is taken for the bed preparation. A slightly different filling procedure is used in this chapter. To prepare the bed before the plate starts moving, particles are first created and put into a container as in Chapter \ref{chapter 2}. The box shaped container's cross section is 40 cm $\times$ 30 cm and it is filled by particles to a depth of approximately 30 cm. Once particles are created and placed, they are given random initial velocities in three directions ($x$, $y$ and $z$). The absolute velocity values for each direction are from 40 cm/s to 45 cm/s and the sign of the velocity is randomly chosen. I wait 2.5 s in order to make sure the granular bed is fully settled before starting the intrusion of the foot. Figure \ref{fig:mean_std} shows the mean speed values of 95700 particles and their standard deviation from 0 to 9.9 s. The logarithmic mean and standard deviation can be found in Figure \ref{fig:log_mean_std}. As the plots show, particle velocities quickly decay to near 0, so 2.5 s is more than enough time to ensure the granular bed is fully settled.   

\section{Resistive forces fitting functions}
\begin{figure}
    \centering
    \includegraphics[width=17cm]{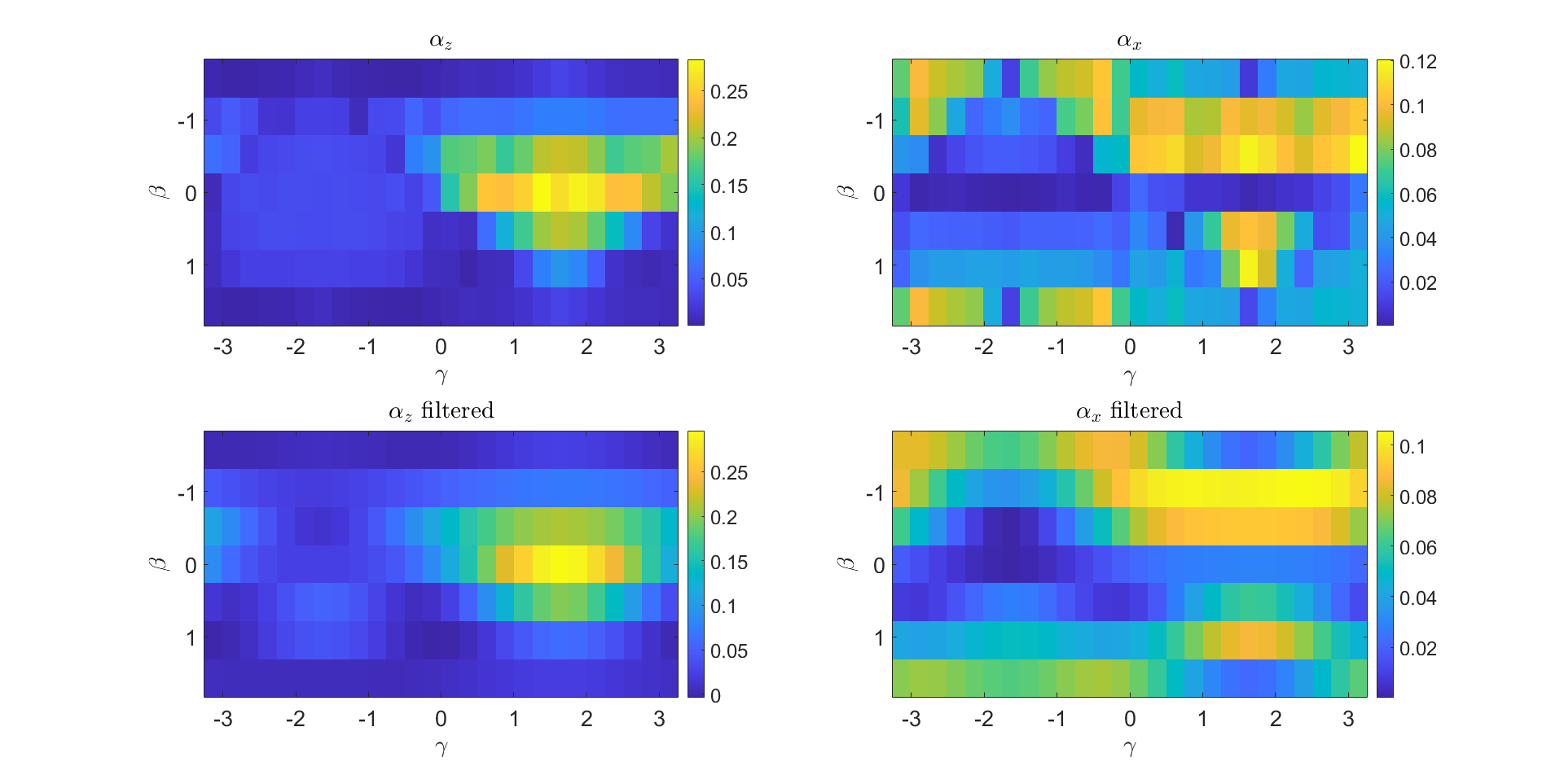}
    \caption[Raw stress gradient colormaps and their Fourier filtered colormaps]{Raw (top row) and Fourier filtered (bottom row) $\alpha_{z,x}$ colormaps ($\gamma$ and $\beta$ values are in radians). Colorbar units are in N/$\textrm{cm}^3$.}
    \label{fig:color_map}
\end{figure}

To get analytic expressions for $\alpha_{x,z}$ in Eq.\ref{stress}, I did hundreds of DEM simulations with different attack angles $\beta$ and intrusion angles $\gamma$ which vary from -90 degrees to 90 degrees. The constant intrusion speed is chosen as 3 cm/s. For each $\beta-\gamma$ pair, I did three repeat trials, and took the average of $\alpha_{x,z}$, as in Chapter \ref{chapter 2}, and then I took absolute values of those $\alpha_{x,z}(\beta,\gamma)$ data. Figure \ref{fig:color_map} plots $|\alpha_{x,z}(\beta,\gamma)|$. In Figure \ref{fig:color_map}, the upper two plots show the $\alpha_z$ and $\alpha_x$ change with different $\beta$ and $\gamma$. I took absolute values of the raw data because the stress results change rapidly when the plate switches from intruding to extracting because the stresses changes from positive to negative. For a small number of modes, the Fourier transform is unable to capture this rapid change. However, since the resistive forces are always opposite to the plate's moving direction, I can easily find the correct resistive forces direction given the plate's velocities. Thus, it's reasonable to take absolute values of the raw data to reduce Fourier fitting errors. By performing discrete Fourier transform on the processed raw $\alpha_{z,x}$ data over $-\pi/2 < \beta < \pi/2$ and $-\pi < \gamma < \pi$, I can obtain a fitting function. To balance the accuracy and conciseness of the fitting formulas, I set the Fourier transform order as 2. The fitting functions are:\newline
\begin{equation}
    \label{alpha_z_function}
    \alpha_{z}(\beta,\gamma) = \sum_{m=-2}^{2}\sum_{n=-2}^{2}[A_{m,n}\cos(2m\beta+n\gamma)+B_{m,n}\sin(2m\beta+n\gamma)] \mathrm{,}
\end{equation}
\begin{equation}
    \label{alpha_x_function}
    \alpha_{x}(\beta,\gamma) = \sum_{m=-2}^{2}\sum_{n=-2}^{2}[C_{m,n}\cos(2m\beta+n\gamma)+D_{m,n}\sin(2m\beta+n\gamma)],
\end{equation}
where $A_{m,n}$, $B_{m,n}$, $C_{m,n}$ and $D_{m,n}$ are coefficients of the fitting function and can be found in Table \ref{Fourier coefficients}. The coefficients for 1st order and 3rd order Fourier fitting can be found in this Google drive file \footnote{\url{https://drive.google.com/file/d/1T4itiIVEqeFl1zM6uPIz8_AqbE3kJiTy/view?usp=sharing}}.
\begin{table}[]
\centering
\begin{tabular}{llllll}
\hline
$m$  & $n$  & $A_{m,n}$         & $B_{m,n}$         & $C_{m,n}$         & $D_{m,n}$         \\ \hline
0  & 0  & 0.0587405 & 0         & 0.0510357 & 0         \\
0  & 1  & -0.000449 & 0.0256221 & 0.0002929 & 0.0062092 \\
0  & 2  & -0.007932 & -0.000897 & 0.0021059 & -0.0005   \\
0  & -2 & -0.007932 & 0.0008966 & 0.0021059 & 0.0004998 \\
0  & -1 & -0.000449 & -0.025622 & 0.0002929 & -0.006209 \\
-1 & 0  & 0.0308869 & -0.0081   & -0.009617 & -0.007194 \\
-1 & 1  & 0.0023856 & 0.0152946 & 0.0026648 & 0.0054052 \\
-1 & 2  & -0.003031 & -0.005611 & -0.002222 & -0.003655 \\
-1 & -2 & -0.00331  & -0.004783 & -0.002184 & -0.003597 \\
-1 & -1 & -0.002843 & -0.016151 & -0.002984 & -0.005976 \\
-2 & 0  & 0.0050729 & -0.00493  & -0.007099 & -0.002165 \\
-2 & 1  & 0.0013374 & 0.0052393 & 0.0015156 & -0.005303 \\
-2 & 2  & -1.17E-05 & -0.003304 & 0.0020161 & -0.000657 \\
-2 & -2 & -0.000354 & -0.003454 & 0.0017459 & -0.001073 \\
-2 & -1 & -0.001479 & -0.005812 & -0.001452 & 0.0051538 \\
2  & 0  & 0.0050729 & 0.0049298 & -0.007099 & 0.0021647 \\
2  & 1  & -0.001479 & 0.0058119 & -0.001452 & -0.005154 \\
2  & 2  & -0.000354 & 0.0034541 & 0.0017459 & 0.0010725 \\
2  & -2 & -1.17E-05 & 0.0033043 & 0.0020161 & 0.0006572 \\
2  & -1 & 0.0013374 & -0.005239 & 0.0015156 & 0.0053032 \\
1  & 0  & 0.0308869 & 0.0080999 & -0.009617 & 0.0071942 \\
1  & 1  & -0.002843 & 0.0161505 & -0.002984 & 0.0059762 \\
1  & 2  & -0.00331  & 0.0047832 & -0.002184 & 0.0035974 \\
1  & -2 & -0.003031 & 0.0056105 & -0.002222 & 0.0036552 \\
1  & -1 & 0.0023856 & -0.015295 & 0.0026648 & -0.005405 \\ \hline
\end{tabular}
\caption{Coefficients for 2nd order Fourier filtered force functions.}
\label{Fourier coefficients}
\end{table}
\newline
\begin{figure}
    \centering
    \includegraphics[width=17cm]{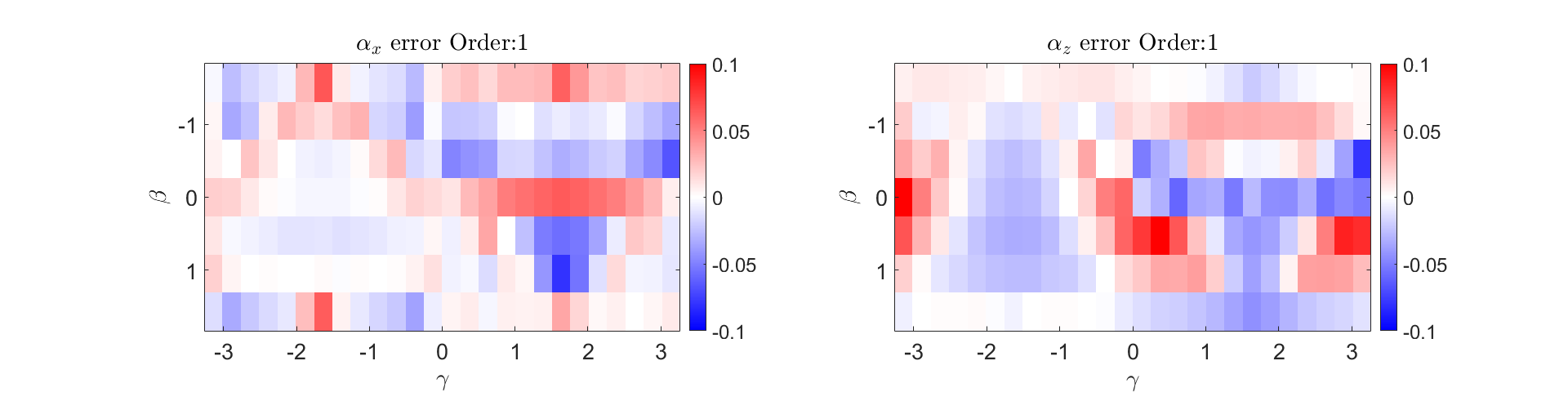}
    \caption[1st order $\alpha_{z,x}$ error maps]{ 1st order $\alpha_{z,x}$ error maps ($\gamma$ and $\beta$ values are in radians). Colorbar units are in N/$\textrm{cm}^3$.}
    \label{fig:error_map_1st}
\end{figure}
\begin{figure}
    \centering
    \includegraphics[width=17cm]{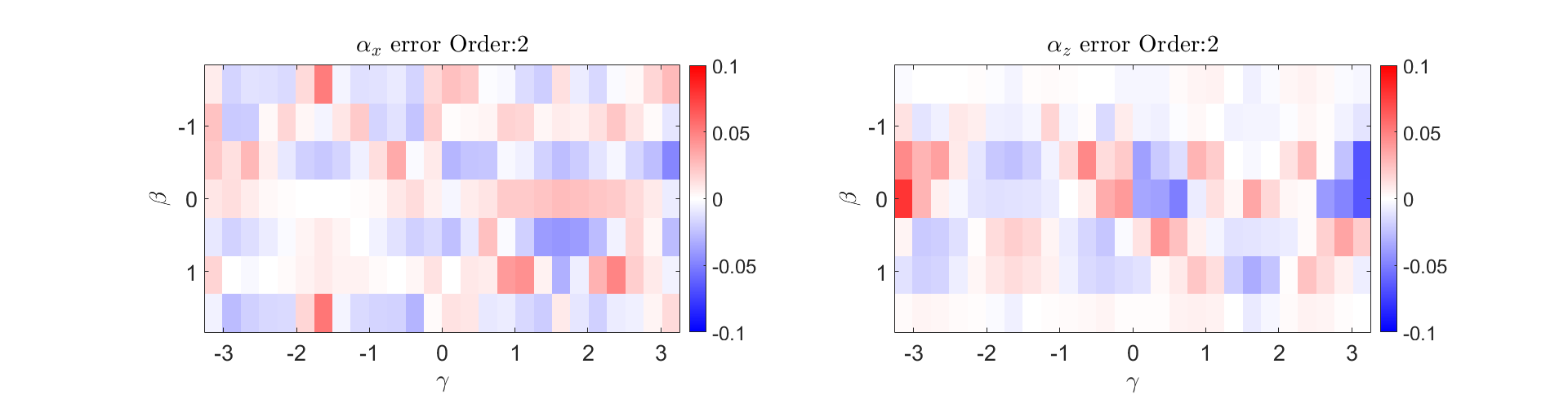}
    \caption[2nd order $\alpha_{z,x}$ error maps]{2nd order $\alpha_{z,x}$ error maps ($\gamma$ and $\beta$ values are in radians). Colorbar units are in N/$\textrm{cm}^3$.}
    \label{fig:error_map_2nd}
\end{figure}
\begin{figure}
    \centering
    \includegraphics[width=17cm]{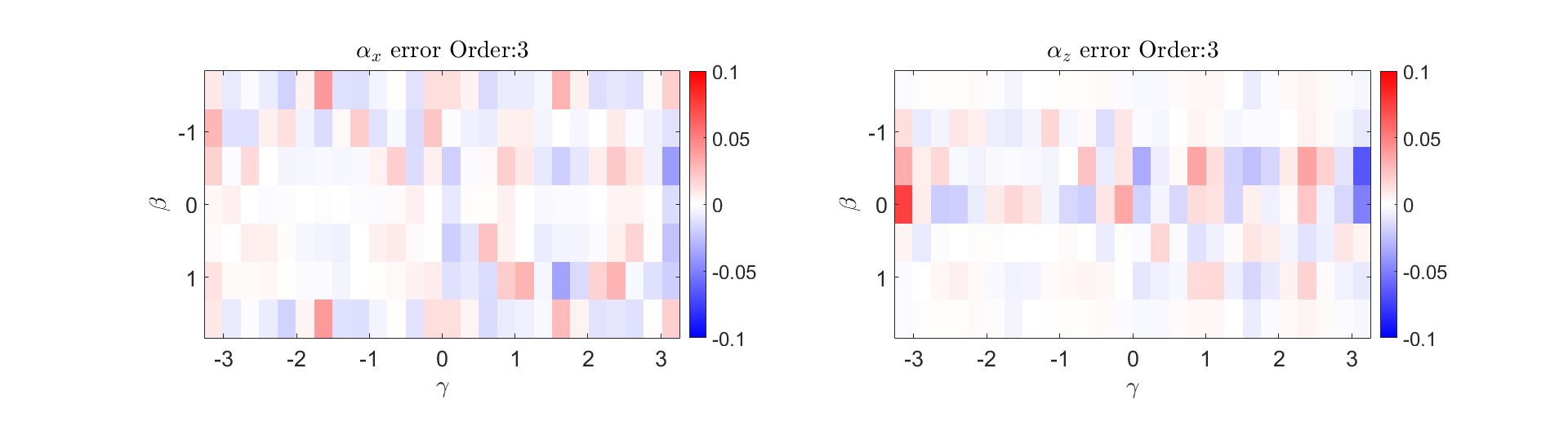}
    \caption[3rd order $\alpha_{z,x}$ error maps]{3rd order $\alpha_{z,x}$ error maps ($\gamma$ and $\beta$ values are in radians). Colorbar units are in N/$\textrm{cm}^3$.}
    \label{fig:error_map_3rd}
\end{figure}
\begin{figure}
    \centering
    \includegraphics[width=17cm]{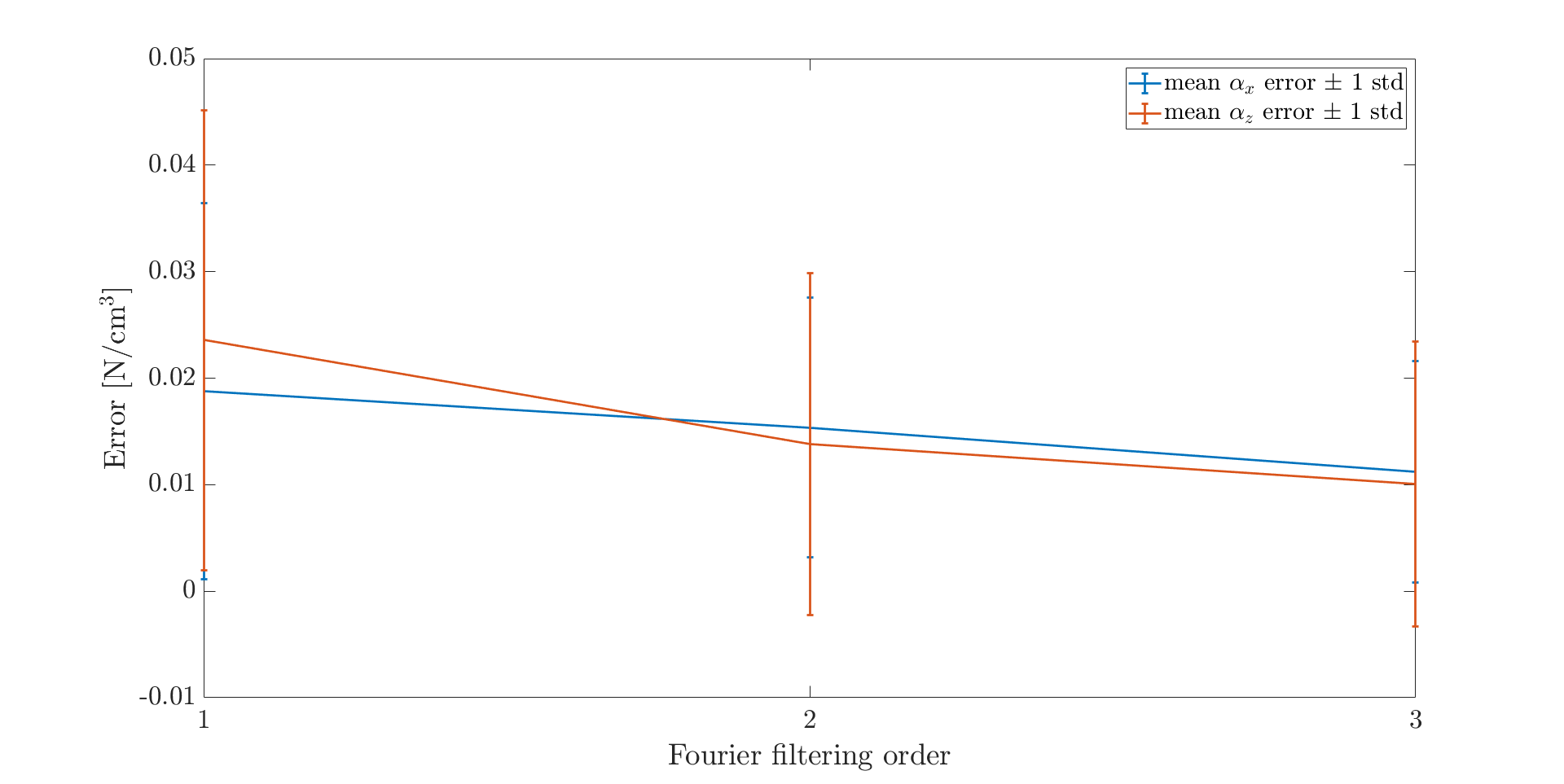}
    \caption[Stress gradient error bars]{Mean error between $\alpha_{x,z}$ from raw data and from Fourier representation vs. number of modes.}
    \label{fig:error_bar}
\end{figure}
To determine the accuracy of the fitting formulas, I generated error maps as shown in Figure \ref{fig:error_map_1st} to Figure \ref{fig:error_map_3rd}. Errors for each $\beta-\gamma$ pair in the error maps are computed as: $\alpha_{z}$ ($\alpha_{x}$) $-$ Fourier fitted $\alpha_{z}$ (Fourier fitted $\alpha_{x}$). By comparing the error maps of three Fourier fitting orders, one can see that a second order fit ensures the fitting formula accuracy while limiting the number of fitting parameters. As Figure \ref{fig:error_map_2nd} shows, errors in most regions are close to zero (less than 0.02) and only errors of $\alpha_{z}$ near $\gamma = \pm\pi$ are near 0.1 $\mathrm{N}/\mathrm{cm}^3$. At the boundaries of the $\beta-\gamma$ plane, the physical meaning is the plate switches between extraction and intrusion. It is more difficult to intrude than withdraw, so  the values near boundaries see a rapid change along $\gamma$. However, low order Fourier fits are less accurate if the raw data changes rapidly. This is why the fit is less accurate near boundaries. Figure \ref{fig:error_bar} shows that the mean and standard deviation of the absolute errors for order 2 is less than 0.0211 $\mathrm{N/cm}^3$, so the 2nd-order Fourier series approximation can be used as a fast-to-evaluate model of granular resistive stresses, at least for low intrusion velocities and away from the effects of boundaries.

%%%%%%%%%%%%%%%%
% Chapter 4
%%%%%%%%%%%%%%%%

\chapter{Periodic Hopping of a Monopod} \label{chapter 4}
Many applications for mobile robots involve environments that are dangerous or unsuitable for humans. Moreover, due to the inaccessible/unsafe nature of these environments, level rigid ground is rare, so wheeled/treaded robots are often unsuitable due to limited mobility. Legged robots would seem the ideal choice, except that the bulk of research and development of legged robots has focused on hard-ground applications, which are of little use in these challenging environments. In many cases, the terrain is not rigid but, rather, is some kind of deformable granular substrate (e.g., sand, soil, or snow). While there has been recent progress on legged robotic locomotion on these kinds of substrates (e.g., \cite{Li_Terradynamics_2013}, \cite{Aguilar2016RobophysicalSO}, \cite{terrain_aware_motion_planning}, \cite{xiong_stability_IROS_2017}), a major remaining challenge with hopping/walking/running on these substrates is that they do not return to their pre-impact state. This represents a permanent energy loss as well as a challenge to maintaining stable gaits. In addition to its relevance to legged locomotion on deformable terrain, this unidirectional ground response represents an interesting variation on the classic ``bouncing ball problem"  in hybrid dynamics \cite{HOLMES_1982},\cite{Aguilar_lift_off_robot}. While there are many tools for generating and analyzing cyclic trajectories in hybrid dynamical systems (e.g., Poincare analysis, Lyapunov analysis), I bring the tools of trajectory optimization \cite{terrain_aware_motion_planning} and discrete-element-method (DEM) simulation to bear on the problem. Specifically, I formulate the search for vertically-constrained hopping gaits on deformable terrain as a trajectory optimization problem with boundary constraints. I solve this problem numerically, obtaining a feedforward control signal. Then, in DEM simulation, I apply this feedforward control signal along with a stabilizing time-invariant feedback law. \newline
In this chapter, I develop a control strategy for a vertically-constrained monopod robot hopping on non-cohesive frictional granular media. The robot is released from the apex of its trajectory and then impacts the bed with velocities ranging from 0 to -4 m/s. When the foot contacts the granular bed, the ground reaction forces are modeled by RFT. Then the equations of motion of the robot has a closed-form solution. The trajectory optimization treats the equations of motion as a part of the equality constraints and produces feedforward control efforts for the robot. Finally, the center of mass (CoM) of the robot can be controlled to return to the initial apex with zero speed after both a single hop and five hops. Both single hop and five hops are tested in DEM using both feedforward and feedforward + feedback control strategies.
\begin{figure}
    \centering
    \includegraphics[width=12cm]{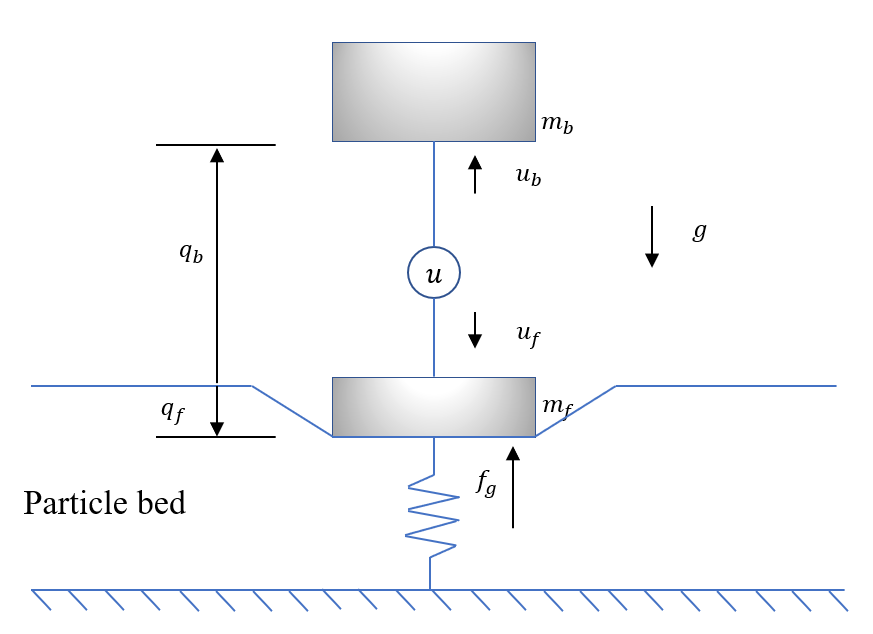}
    \caption[Physical models of the monopod robot and soft ground.]{Physical models of the monopod robot and soft ground. The soft ground is treated as a unidirectional spring with stiffness $k_g$, see Eq. \ref{GRF}. The body, located at height $q_b$ above the undisturbed ground, has mass $m_b$, and the foot, located at height -$q_f$ below the undisturbed ground, has mass $m_f$. Body and foot are connected with a linear motor which exerts a possibly time dependent and equal and opposite force $u$ on both masses. Heights $q_b$ and -$q_f$ are measured relative to the undisturbed ground surface. }
    \label{fig:monopod}
\end{figure}
\section{Soft ground model}
Eq. \ref{stress} shows that the horizontal and vertical resistive forces increase linearly with the intrusion depth. For a 1D hopping monopod robot, the granular substrate can be modeled as a unidirectional spring. The ground resistive force $f_g$ is expressed as
\begin{equation}
    \label{GRF}
f_g = \left\{
\begin{array}{ll}
0 & \textbf{flight}\\
-k_{g}q_{f} &  \textbf{yielding}\\
\textbf{[}0,-k_{g}q_{f}\textbf{]} & \textbf{static}
\end{array} \right.
\end{equation}
where $k_g$ is the ground stiffness. The area of the foot is 25 $\textrm{cm}^2$ and the stress gradient $\alpha_z(\beta = 0,\gamma = 90 ^{\circ}$) is 0.25 $\textrm{N/cm}^3$. I chose $k_g =25\textrm{ cm}^2\times 0.25\textrm{ N/cm}^3 \times 100 = 625 \mathrm{N/m}$ to match the experimental data in Chapter \ref{Chapter 3} I ignore the withdraw force when lift off the robot. The foot thickness is chosen to ensure no particles can accumulate on the top of the foot, which helps to reduce the withdraw stresses.
\section{Robot model}
\begin{figure}
    \centering
    \includegraphics[width =17cm]{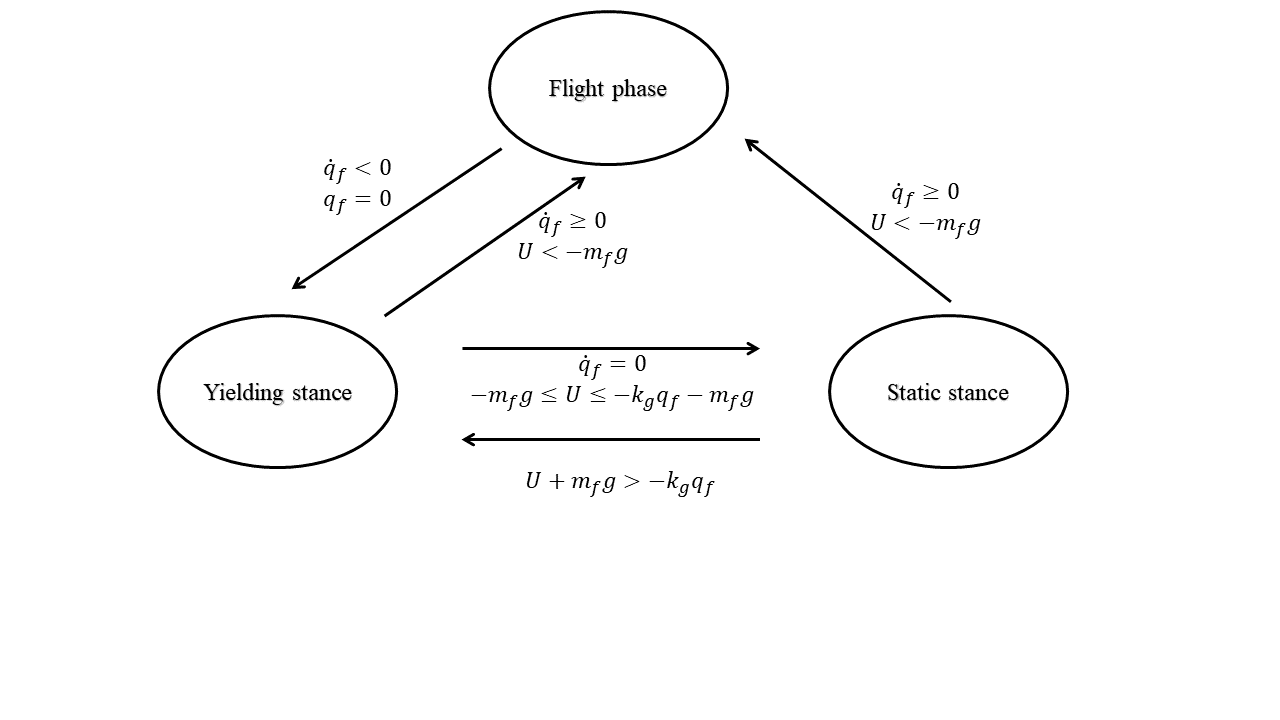}
    \caption[Sketch of phase transitions]{Transitions between flight phase, yielding stance and static stance. Reproduced from \cite{soft_landing} with permission.}
    \label{fig:phase_transition}
\end{figure}

To focus on foot-ground interaction rather than whole-body control, I study a simple monopedal robot hopping vertically as illustrated in Figure \ref{fig:monopod}. The robot consists of a body (position $q_{b}$ and mass $m_{b}$) and a flat-bottomed foot (position $q_{f}$ and mass $m_{f}$) connected by a linear motor (idealized as a source of force $u$) such that $u > 0$ pushes the two masses apart. The dimensions and masses of the robot foot and body can be found in Table \ref{table:monopod}. The stroke limits of the robot system are
\begin{equation}
    \label{stroke limit}
    0 < q_b - q_f < 0.5\ \mathrm{m.}
\end{equation}
\begin{table}[]
\centering
\begin{tabular}{ll}
\hline
Foot dimension & 5 cm x 5 cm x 10 cm \\
Foot mass      & 0.25 kg           \\
Body dimension & 5 cm x 5 cm x 5 cm   \\
Body mass      & 1.25 kg            \\ 
Stroke limit  & 0.5 m               \\\hline
\end{tabular}
\caption{Physical properties of the monopod robot.}
\label{table:monopod}
\end{table}
The hybrid dynamics of the hopping robot are divided into three phases: flight, yielding stance, and static stance. Figure \ref{fig:phase_transition}, reproduced from~\cite{soft_landing}, describes the conditions for transitioning between flight, yielding stance, and static stance. Phase transitions depend on the foot kinematics and the control effort. My robotic task is represented by periodicity constraints on the CoM position and velocity, but during flight, I have no control over the CoM trajectory. Thus, any attempt to influence the flight-phase CoM trajectory must happen during stance. Therefore, I restrict my analysis to the static and yielding stance phases. The equations of motion for the two stance phases are,
\begin{equation}
\label{stance_1}
\ddot{q_b} = \frac{u}{m_{b}} -g \mathrm{ , and}
\end{equation}
\begin{equation}
\label{stance_2}
\ddot{q_f} = \frac{f_{g}}{m_f}-\frac{u}{m_f}-g \mathrm{ ,}
\end{equation}
where $g$ is the gravitational acceleration ($9.81 \ \mathrm{m}/\mathrm{s}^2$). 
For the following section on robot control, initial conditions are,
\begin{equation}
    \label{initial conditions}
    \begin{array}{c}
      q_b(0) = 0.25\textrm{ m, } q_f(0) = 0 \textrm{ and } \dot{q}_b = \dot{q}_f = V_{0} \textrm{,} 
    \end{array}
\end{equation}
where $V_0$ is the impact velocity.
\section{Robot control}
My first task is to find period one hopping gaits for the vertically-constrained monopod robot. Gait generation is the formulation and selection of a sequence of coordinated leg and body motions that propel a legged robot along a desired path \cite{David_1992_gait_generation}. Specifically, my robot starts impact at a certain initial position and moves vertically downward with an initial impact velocity. The desired motor control periodically returns the robot to the initial impact position with the same CoM velocity magnitude but in the upward direction on each successive hop. \newline
There are various methods of determining the control signal $u(t)$ that results in periodic hopping trajectories (e.g., Poincare analysis \cite{westervelt2018feedback}, Lyapunov analysis), but I use trajectory optimization \cite{OptimTraj} to find controls and trajectories that satisfy periodicity constraints.

\subsection{Feedforwrd control}
\begin{figure}
    \centering
    \includegraphics[width=17cm]{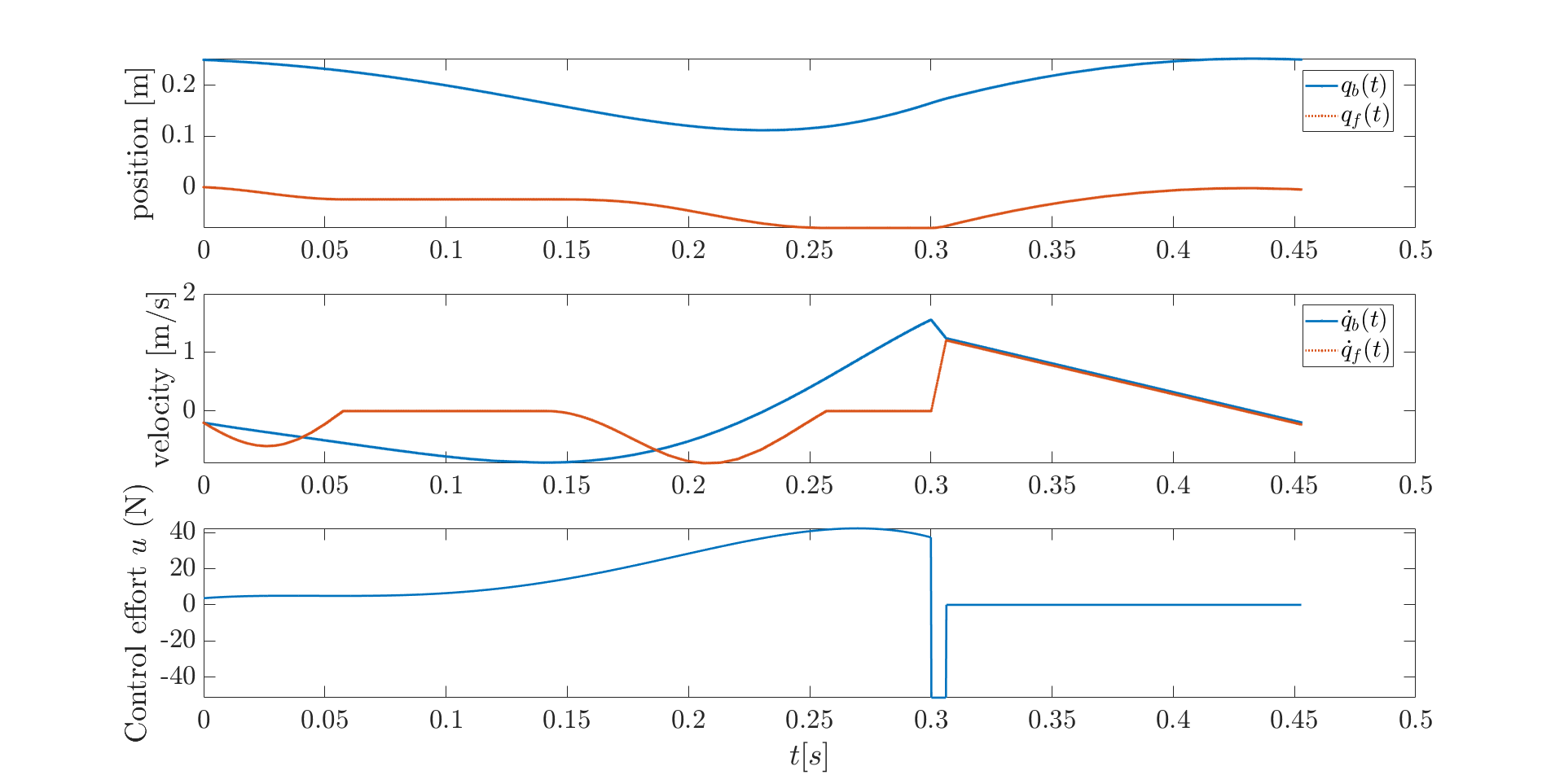}
    \caption[Planned single-hop monopod trajectories and their control signals]{Planned foot and body states vs.\ time (upper and middle plots) and planned control efforts vs.\ time (bottom plot).}
    \label{fig:foot and body and control signals}
\end{figure}

To determine the open loop control signal for the robot, I divide the problem into two parts, stance phase and flight phase. The stance-phase optimal control problem can be transformed into a constrained nonlinear program. For the flight phase, no ground forces are applied to the foot, the velocity of body and foot is expected to be equal, and the distance between the body and foot should return to the initial value, so analytic methods are used to determine the corresponding control signal.\newline
Hubicki \etal $\ $\cite{terrain_aware_motion_planning} describe a fast optimal motion planning algorithm for a 1D hopping robot. By formulating an optimal control problem as a constrained nonlinear program (NLP), I utilize a well-developed NLP solver (i.e., MATLAB fmincon) to find out the control signal. Control signals $u(t)$ are chosen as fourth order polynomials because the system has 4 constraints, which determine 4 coefficients of the polynomials. The remaining coefficient is determined by optimizing the cost function. The expression is as follows,
\begin{equation}
    \label{u(t)}
    u(t) = a + bt + ct^2 + dt^3 + et^4 \mathrm{,}
\end{equation}
where $\mathbf{x} = [a,b,c,d,e]$ are the design variables of the NLP. Note that I fix the time at 0.3 s for the stance phase and optimize the coefficients of the control signals $u(t)$ to satisfy the constraints. I choose 0.3 s because it can ensure my MATLAB code to find a solution for the optimal control. For a longer time like 1s can also make the optimization problem converge but it needs more time to solve. In addition, too short a stance time will not be able to develop enough momentum for the robot to jump back to apex given stroke constraints and lack of rate dependence in GRF model.\newline
I choose control effort squared as the cost function:
\begin{equation}
    \label{cost_func}
    J(\mathbf{x}) = \int_{0} ^ {T_s} u^{2}(t) dt \textrm{,}
\end{equation}
Reference \cite{soft_landing} chooses intrusion depth of the foot as the cost function. I choose the control effort squared because it makes the cost function convex which is good for the convergence of the optimization problem.
The resulting formation for the stance-phase nonlinear program is as follows:

\begin{equation}
    \label{stance_NLP}
    \begin{array}{lll}
  \mathbf{x}^{*} &=\displaystyle \argmin_{\mathbf{x}} J(\mathbf{x})    &\textrm{s.t.}  \\
         f(\mathbf{x}) &= 0 \textrm{,}\\
         g(\mathbf{x}) &\leq 0\mathrm{,}
    \end{array}
\end{equation}
where $f(\mathbf{x})$ are the equality constraints obtained by the robot dynamics (see Eq. \ref{stance_1} and Eq. \ref{stance_2} ) and robot boundary conditions, $g(\mathbf{x})$ are the inequality constraints obtained by the linear controller control limits and robot stroke limits (see \ref{stroke limit}), and $T_s$ is the desired stance-phase period.\newline
The stance-phase terminal conditions at $T_s$ are
\begin{equation}
    \label{terminal conditions}
    \begin{array}{cc}
    q_b(T_s) = C\textrm{, }q_f(T_s) =D\textrm{, }\dot{q}_b(T_s) = E      \textrm{ and } \dot{q}_f(T_s) = 0  \mathrm{,}
    \end{array}
\end{equation}
where $[C$\textrm{, }$ D$\textrm{, }$E]$ (terminal body position, terminal foot position, and terminal body velocity respectively) are constants determined by ground stiffness and impact velocity. One thing interesting is my robot never leaves ground in the stance phase because I add constraints in the MATLAB solver to do this.\newline
 The flight phase control first applies a constant forces to the foot and body, and then turns off the linear motor during the remainder of free fall. I choose this strategy for the flight phase because the solution for control efforts is easily obtained. Once given the depth of the foot at the end of the stance phase and the lift-off acceleration of the foot, I can solve for the control efforts for lift-off and the terminal states of the body in the stance phase.\newline
By setting the stance-phase duration to 0.3 s and boundary conditions as given in Table \ref{table:BC}, I obtain the control efforts and the resulting body and foot trajectories shown in Figure \ref{fig:foot and body and control signals}. As those results are for an example impact velocity of -0.2 m/s, the foot and the body move downwards from the top of the granular media. The control signals make the foot and the body return to the initial height while the terminal speed values are the same as the initial speed. We can also observe the double intrusion of the foot, which is an interesting feature of locomoting on soft ground. \newline
\newline
\begin{table}[]
\centering
\begin{tabular}{lllr}
\hline
$q_f(0)$ & 0 m    &  $\dot{q}_f(0)$& -0.2\ m/s \\
$q_b(0)$ & 0.25 m &  $\dot{q}_b(0)$& -0.2\ m/s \\
$q_f(T)$ & 0 m    &  $\dot{q}_f(T)$& 0.2\ m/s  \\
$q_b(T)$ & 0.25 m &  $\dot{q}_b(T)$& 0.2\ m/s  \\ \hline
\end{tabular}
\caption{Boundary conditions}
\label{table:BC}
\end{table}
\subsection{Feedforward plus feedback control} \label{FpF control}
\begin{figure}
    \centering
    \includegraphics[width=17cm]{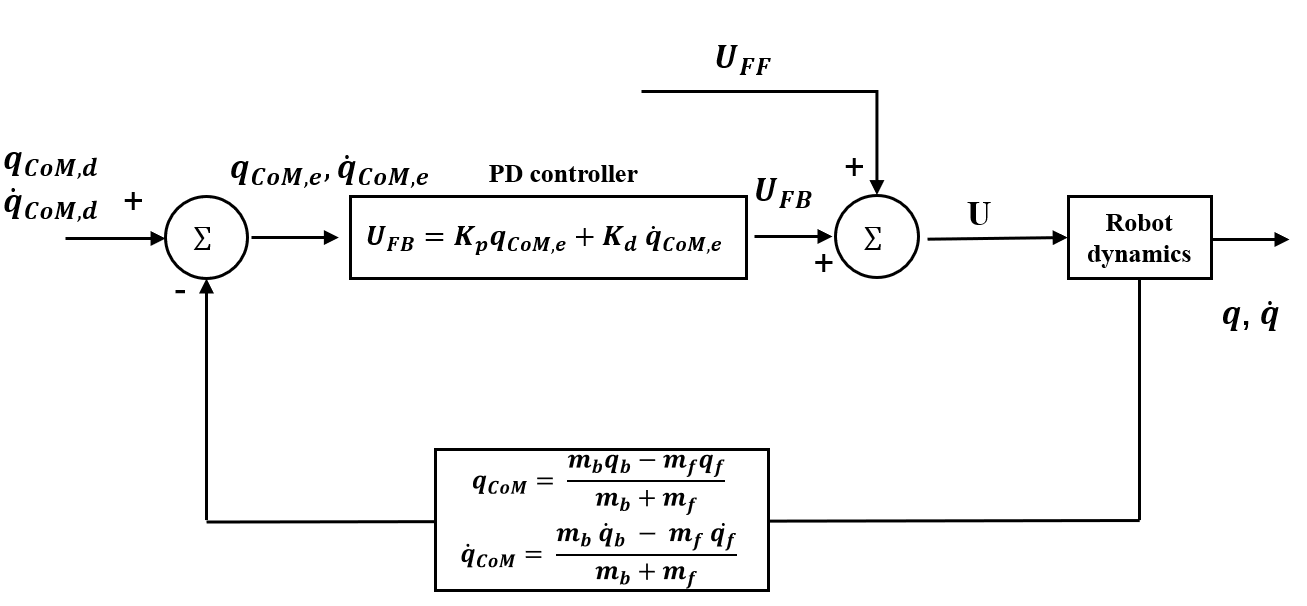}
    \caption[Block diagram of feedforward plus feedback control for stance phase]{Block diagram of feedforward plus feedback control for stance phase. $q_{CoM}$ and $\dot{q}_{CoM}$ are the center of mass position and velocity, respectively. $q_{CoM}$, $q_{CoM,d}$, and $q_{CoM,e}$ are actual center of mass position, desired position and position error, respectively. $U_{FB}$ and $U_{FF}$ are the feedback and feedforward control forces, respectively.}
    \label{fig:block_diagram}
\end{figure}

\begin{figure}
    \centering
    \includegraphics[width=17cm]{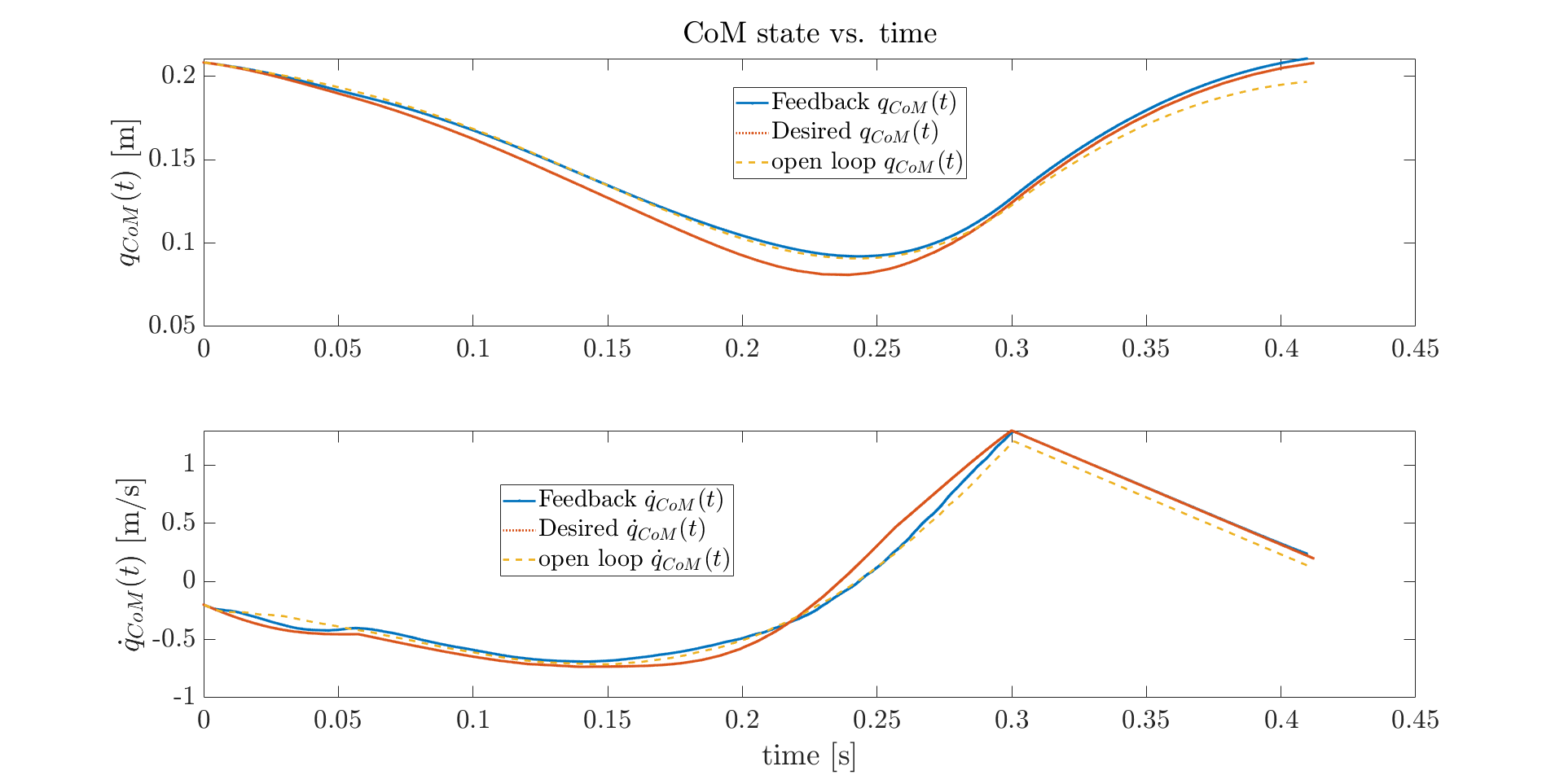}
    \caption[CoM trajectories from optimal(desired) and for the two different control strategies]{CoM trajectories from optimal(desired) and for two control strategies. Upper plot: CoM position vs. time. Bottom plot: CoM velocity vs. time. Blue, red and yellow curves are feedforward plus feedback results, expected results from MATLAB and feedforward control results, respectively.}
    \label{fig:CoM state}
\end{figure}

\begin{figure}
    \centering
    \includegraphics[width=17cm]{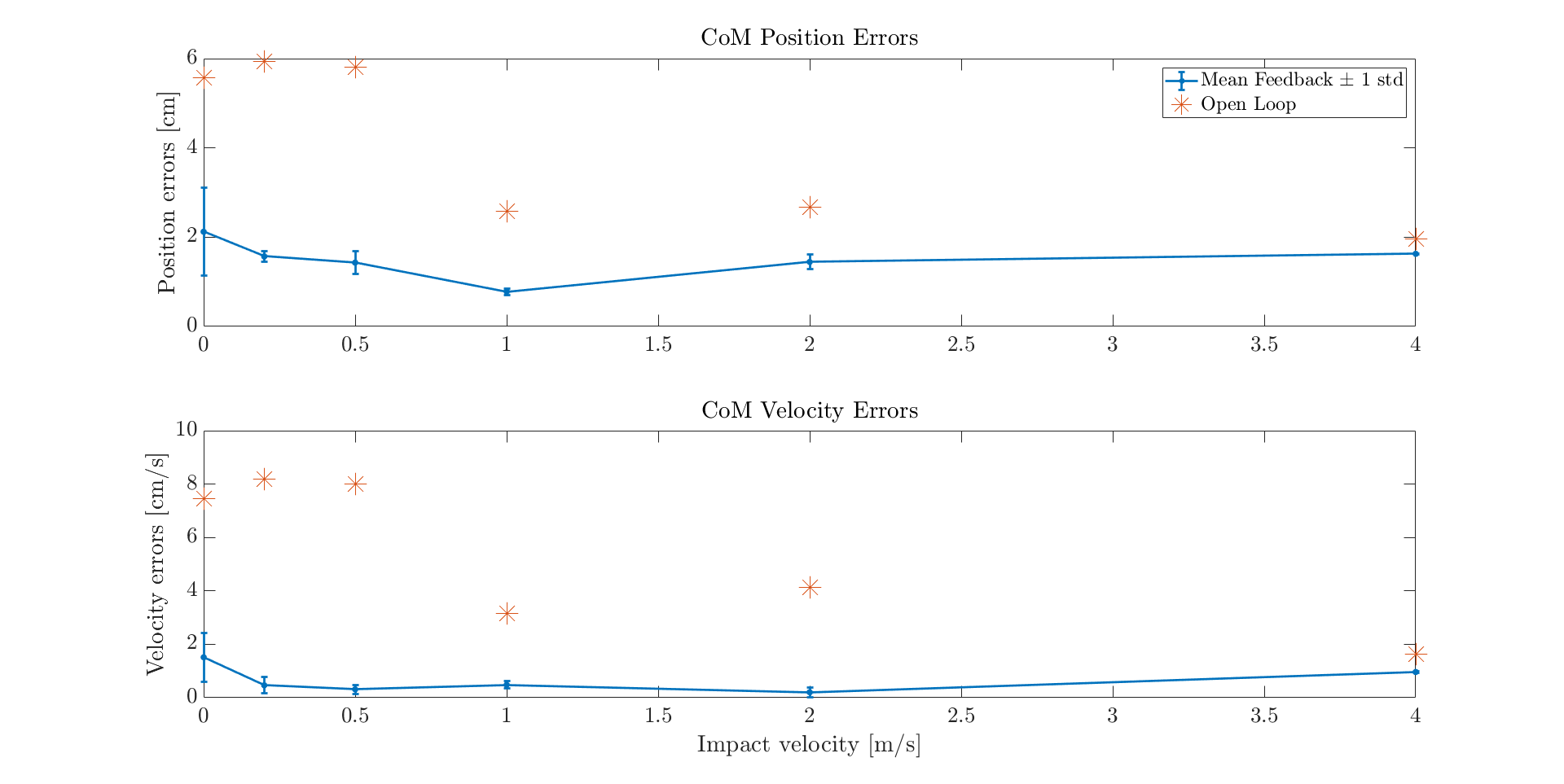}
    \caption[Position and velocity errors of two different control strategies for multiple impact velocities]{Position and velocity errors of two different control strategies for multiple impact velocities. Asterisks are errors of open loop control while error bars are errors of feedforward plus feedback control.}
    \label{fig:total_error}
\end{figure}

\begin{figure}
    \centering
    \includegraphics[width=17cm]{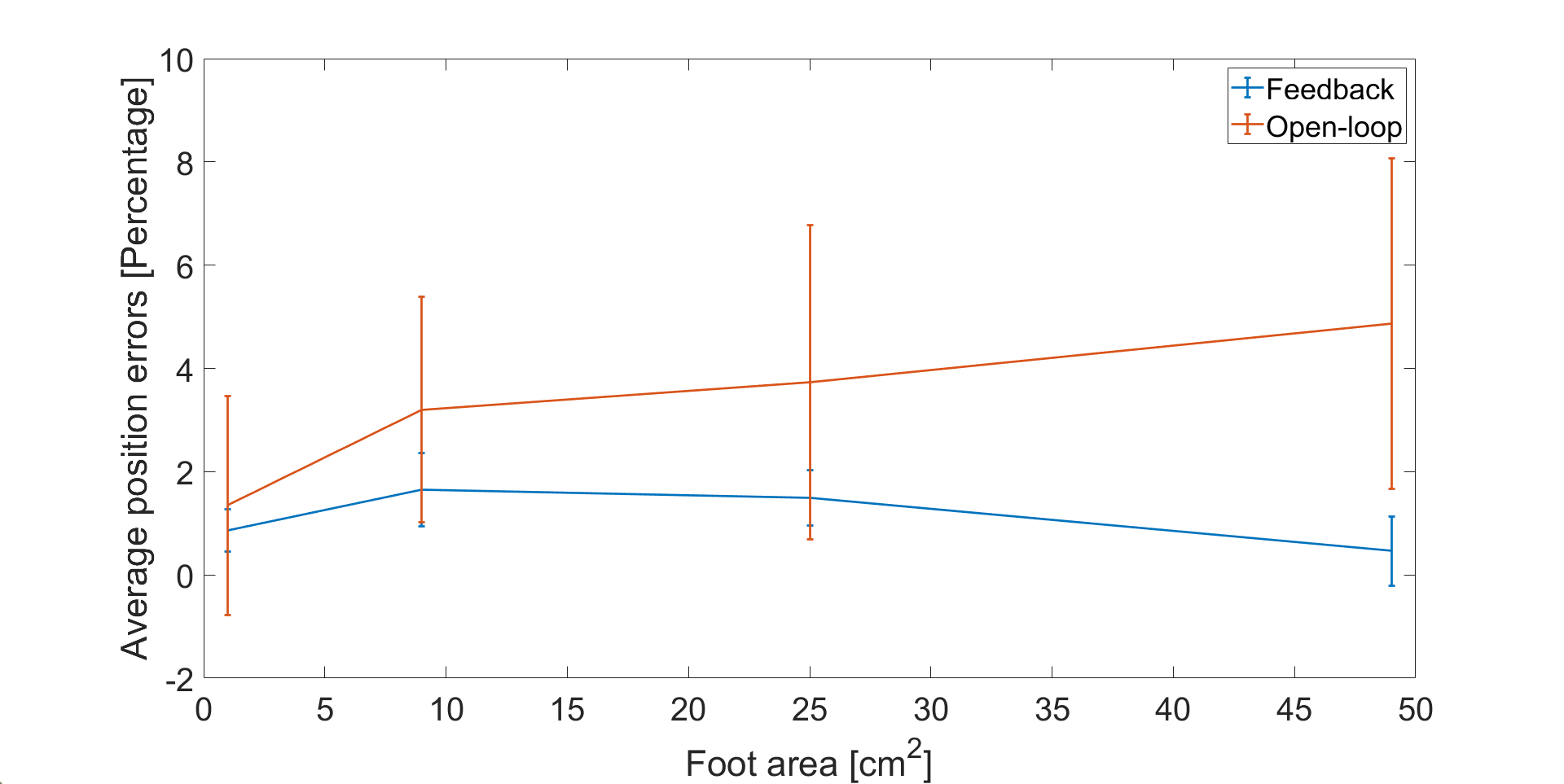}
    \caption[Experimental errors vs.\ foot size]{Experimental errors in CoM position from Eq.\ref{error_percent} vs. foot size.}
    \label{fig:error_overview}
\end{figure}

After testing the open loop control forces in Chrono simulation, I found that the trajectories of the robot deviate from the expected ones because the ground stiffness approximation is not sufficiently accurate. Therefore, I introduce a feedback controller to stabilize control in stance phase. So, the control strategy is feedforward plus feedback control now. \newline
As Figure \ref{fig:block_diagram} shows, the control system inputs are the desired center of mass(CoM) position and velocity of the robot. The desired CoM states($q_{CoM,d}\textrm{ and }\dot{q}_{CoM,d}$) are generated from MATLAB when I apply the open loop control forces to the robot in numerical simulation. The difference between the desired robot states and the actual robot states ($q_{CoM}\textrm{ and }\dot{q}_{CoM}$) are the CoM errors($q_{CoM,e}\textrm{ and }\dot{q}_{CoM,e}$). By passing those errors into a PD controller, I obtain the feedback control forces $U_{FB}$ as illustrated in Fig.\ref{fig:block_diagram}. In the section above, I use trajectory optimization to get the feedforward control forces($U_{FF}$). Combining the feedback and feedforward control forces and adding the new forces($U$) into the equations of motion of the robot system, I get the real-time states of the robot.\newline 
I implement this feedforward plus feedback controller on my robot model in Chrono. Specifically, I create a robot model like Figure \ref{fig:monopod} with physical properties as given in Table \ref{table:monopod}. I set the impact velocity of both the body and the foot to -0.2 m/s. When the robot transitions from stance phase to flight phase, the CoM position is expected to return to 0, and the CoM velocity is expected to be 0.2 m/s. Since no external force is applied to the robot system during the flight phase, the robot can return to the initial position with the same velocity because of momentum conservation. The two conditions thus ensure the robot hops periodically. For the open loop control case, I applied the feedforward control forces generated from MATLAB. For the feedforward plus feedback case, I add a PD controller on the robot to track the CoM of the robot. As Figure \ref{fig:CoM state} shows, my controller significantly reduce both the CoM position and velocity errors.\newline
To quantitatively analyze the error of the two control strategies (i.e.\ feedforward and feedforward+feedback), I test their performance for different impact velocities(from -4 m/s to -0). The errors are computed as the terminal CoM position/velocity of the two control methods in Chrono minus the desired terminal CoM position/velocity from MATLAB. For the feedforward plus feedback control, I run Chrono simulation for 3 trials to analyze its uncertainty. As Figure \ref{fig:total_error} shows, my strategy works well in reducing the error for different impact velocities compared to the open loop control alone.\newline
In addition, I investigate how different foot sizes influence the performance of my controllers. As the foot size decreases toward the particle size, force fluctuations increase and present greater challenges to the controllers. To investigate the efficacy of the controller, I test three foot sizes (1 cm x 1 cm, etc.) in addition to the original 5 cm x 5 cm foot. To ensure the foot reach the same depth for different foot sizes, I scale the mass of the foot and the body based on the nondimensionalization method in \cite{soft_landing}. The ratio of the total mass of the robot to the foot square area is fixed. For each foot size, I test six different impact velocities(0, -0.2m/s, -0.5m/s, -1m/s, -2m/s and -4m/s) and conduct three trials for each impact velocity. Both the feedforward + feedback controller and the open loop controller are tested in these experiments. The CoM position error percentage is computed as Eqs. \ref{error_percent},
\begin{equation}
    \label{error_percent}
    \begin{array}{ll}
    \eta& = \frac{q_{CoM,i}(T)-q_{CoM,d}(T)}{q_{CoM,d}(T)}\times100\%,
    \end{array}
\end{equation}
where $i$ can be feedforward + feedback or feedforward alone, $d$ is the desired CoM position generated from MATLAB, and $T$ is the duration of one hop. For each foot size, I compute the mean values and standard deviations of all the error percentages of all impact velocities, and draw errorbars as shown in Figure \ref{fig:error_overview}. Again, I observe that the feedforward plus feedback controller reduces the error to less than 2$\%$ in all cases. This indicates my controller is robust for different foot sizes, even in the presence of large variations in the ground reaction force when the foot is small. 
\section{Multiple hops}
\begin{figure}
    \centering
    \includegraphics[width=17cm]{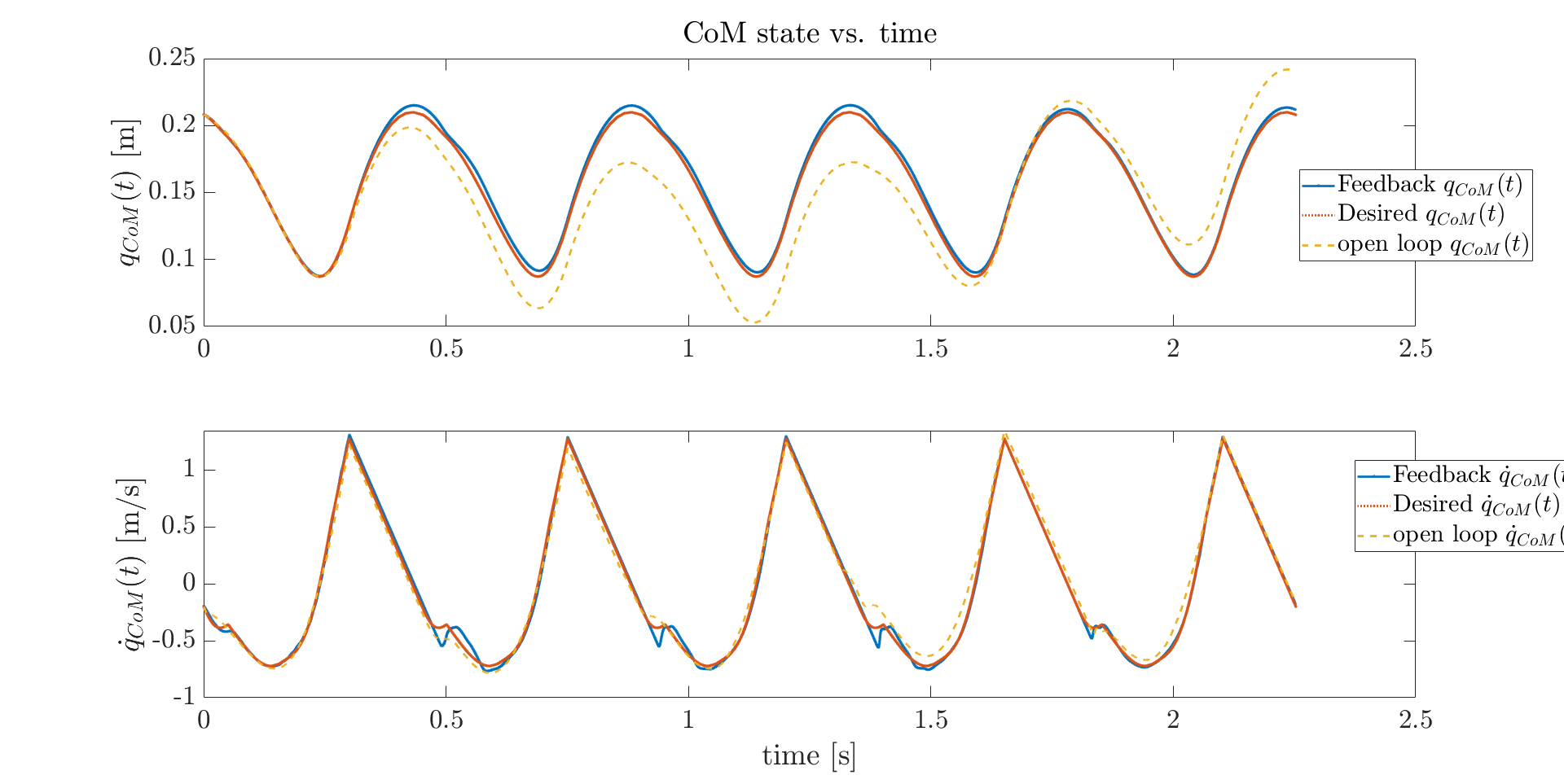}
    \caption[Center of mass trajectories for five hops]{Center of mass trajectories for five hops. Upper plot: CoM position vs. time. Bottom plot: CoM velocity vs. time. Blue, red and yellow curves are feedforward plus feedback results from Chrono, expected results from MATLAB, and feedforward control results from Chrono, respectively.}
    \label{fig:multi_CoM}
\end{figure}

\begin{figure}
    \centering
    \includegraphics[width=17cm]{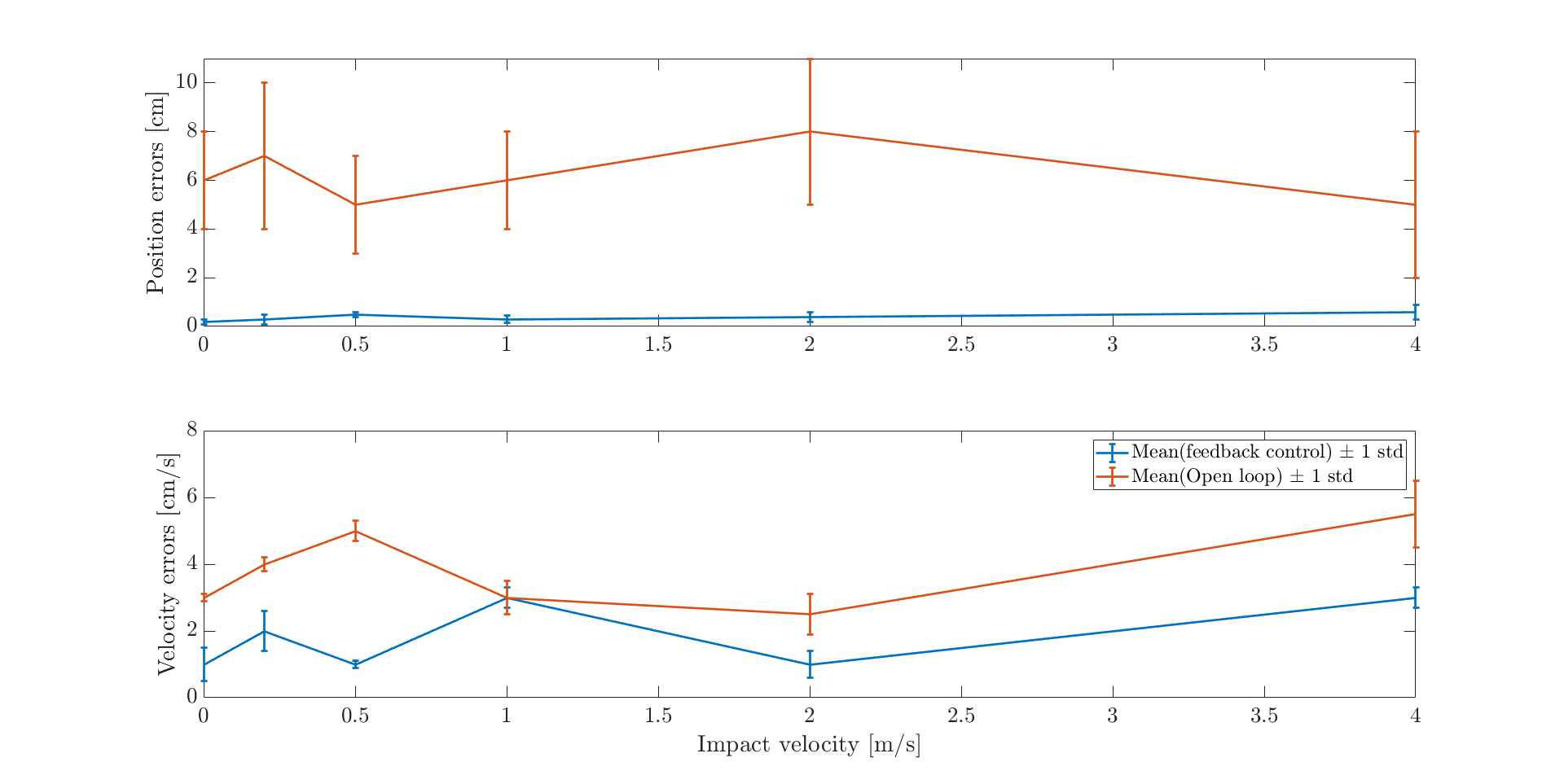}
    \caption[Error of robot CoM trajectories after five hops for different impact velocities]{Errorbar of robot CoM trajectories after five hops for different impact velocities under open loop (orange) and feedforward plus feedback (blue) control.}
    \label{fig:errorbar_5hops}
\end{figure}

\begin{figure}
    \centering
    \includegraphics[width = 17cm]{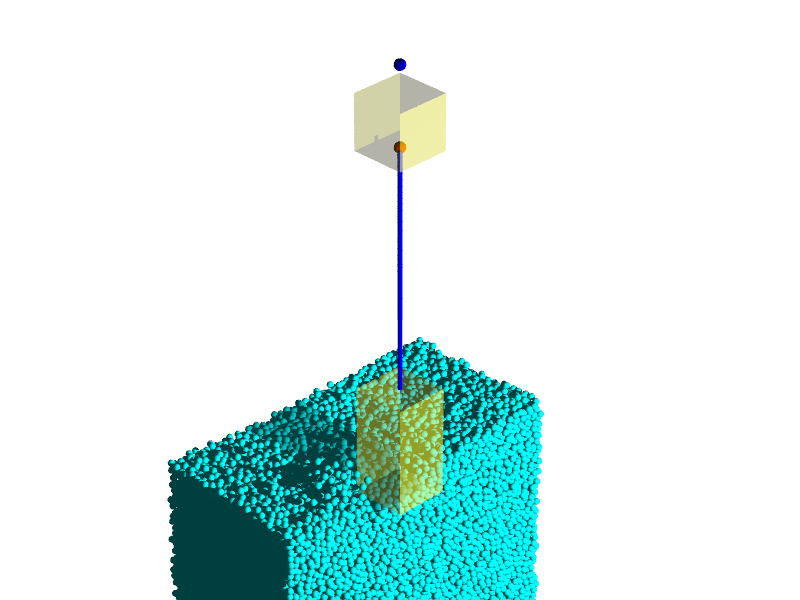}
    \caption[Screenshot of demo video]{Screenshot of demo video. Upper yellow box is the body while bottom yellow box is the foot. The blue line represents the linear motor which connects the body and the foot. Red sphere is the real time position of the CoM of the robot. The blue sphere is the desired apex for the CoM.}
    \label{fig:monodpod_screenshot}
\end{figure}

The ultimate goal of my work is to design gaits for monopod hopping on deformable ground. For the single hop in the sections above, the robot jumps on undisturbed ground. To challenge my feedforward + feedback controller, I make the monopod to have five hops on disturbed ground. As Section \ref{FpF control} shows, I successfully implemented my feedforward plus feedback controller on the robot for single hop. Figure \ref{fig:foot and body and control signals} shows expected trajectories of a single hop of the monopod. The robot starts hopping with impact velocities of -0.2 m/s and returns to the initial position with the same velocity. My control strategy for mutiple hops is to track the CoM trajectories of the single hop of the robot in each hop. In each hop of a multi-hop sequence, I set the period equal to the duration of expected trajectories from MATLAB and use my feedforward plus feedback controller to stabilize my control in stance phase. For the flight phase, I add a PD controller to make the velocity of the body and foot equal, and make the body-foot distance equal to the initial value. Figure \ref{fig:multi_CoM} compares the CoM trajectories of two control strategies. The robot starts moving with an impact velocity of -0.2 m/s. The feedforward plus feedback controller has far greater performance than the open loop controller. After five hops, the CoM trajectories of the robot tracks the expected trajectories quite well. \newline
To compare the errors associated with the two control strategies, I compute the errors as CoM position/velocity at the end of five hops minus the expected terminal CoM position/velocity. To ensure my controller is robust for different impact velocities, I test impact velocities ranging from -4 m/s to 0. For each impact velocity, I run Chrono simulation for three trials. Figure \ref{fig:errorbar_5hops} shows errorbars of both CoM position and CoM velocity after five hops for each impact velocity. The errors of the feedforward plus feedback controller are quite close to zero for all the impact velocities I tested. This again shows the robustness of my controller. In addition to the plots in this chapter, I also made corresponding videos of the robot hopping in Chrono. The first video\footnote{\url{https://drive.google.com/file/d/1LkUWFWx_U1X0XtGNpdf6Dcj_IhaWzxXH/view?usp=sharing}} shows five hops on soft ground using a feedforward plus feedback controller. The robot in the second video\footnote{\url{https://drive.google.com/file/d/1E8iMqUuT_bcmHaP9IYCvcwEjnf4R89iO/view?usp=sharing}} is controlled by an open loop controller. Figure \ref{fig:monodpod_screenshot} is the screenshot of one demo video. Comparing the two videos shows that the feedback controller works well making the CoM to reach the expected height.

%%%%%%%%%%%%%%%%
% Chapter 5
%%%%%%%%%%%%%%%%

\chapter{Conclusion}

In this work, I set up a DEM based simulation environment called Chrono. To validate its reliability and accuracy, I ran many simulations and compared my results with results in \cite{Li_Terradynamics_2013}. The simulation results have a good match to experimental results. After that, I utilized Chrono to generate a ground stiffness approximation formula based the ground resistive force model in \cite{Li_Terradynamics_2013}. I also formulated a soft-landing problem and develop solutions to design open loop periodic gait for our monopod hopper with the trajectory optimization and  the approximation formula. Then I develop a new control solution by applying feedforward plus feedback control on the monopod robot. The new control strategy is tested on single hops and multiple hops for various impact velocities in Chrono simulation. The Chrono simulation results indicate our controller is robust and can reduce CoM trajectory errors. 

This work proves controlling a monopod to desired height periodically when  hopping on yielding terrain is possible. However, our solution for the control force is not optimal, at least in terms of the energy lost to ground. My feature plan is to seek the optimal control that ensures the periodically hopping while minimizes the intrusion depth. Besides, all the experimental results  I got are based on DEM simulation. It's interesting to carry out experiment validation on a real robot. 

%%%%%%%%%%%%%%%%
% References
%%%%%%%%%%%%%%%%

\begin{singlespace}  % use single-line spacing for multi-line text within a single reference
	\setlength\bibitemsep{\baselineskip}  %manually set separataion betwen items in bibliography to double space
	\printbibliography[title={References}]
\end{singlespace}

\addcontentsline{toc}{chapter}{References}  %add References section to Table of Contents

%%%%%%%%%%%%%%%%
% Appendices
%%%%%%%%%%%%%%%%

%\input{appendix.tex}

%%%%%%%%%%%%%%%%
% Vita 
% Only for PhD students
% Masters students remove this line
%%%%%%%%%%%%%%%%

\end{document}